\documentclass[lettersize,journal]{IEEEtran}
\usepackage{amsmath,amsfonts}
\usepackage{algorithmic}
\usepackage{algorithm}
\usepackage{array}
\usepackage[pagebackref,breaklinks,colorlinks,citecolor=black,linkcolor=black, urlcolor=black]{hyperref}
\usepackage[caption=false,font=normalsize,labelfont=sf,textfont=sf]{subfig}
\usepackage{textcomp}
\usepackage{stfloats}
\usepackage{url}
\usepackage{verbatim}
\usepackage{graphicx}
\usepackage{cite}
\usepackage{bm}
\usepackage{booktabs}
\usepackage{multirow}
\usepackage{enumerate}
\begin{document}

\title{Self-Assessed Generation: Trustworthy Label Generation for Optical Flow and Stereo Matching in Real-world}

	\author{Han~Ling,
	Yinghui~Sun,
	Quansen~Sun,
	Ivor~Tsang,
	and
	Yuhui Zheng
	\IEEEcompsocitemizethanks{\IEEEcompsocthanksitem Han Ling and Quansen Sun are with the School of Nanjing University of Science and Technology, Nanjing, Jiangsu; Yinghui Sun is with the School of Southeast University; Ivor Tsang is with the Center for Frontier AI Research, Agency for Science, Technology and Research. Yuhui Zheng is with the Nanjing University of Information Science and Technology. (Corresponding authors: Quansen Sun) \protect\\
		E-mail: 321106010190@njust.edu.cn, sunyh@seu.edu.cn,
		sunquansen@njust.edu.cn, ivor.tsang@gmail.com, zhengyh@vip.126.com}   


\thanks{Manuscript received April 19, 2021; revised August 16, 2021.}}

\markboth{Journal of \LaTeX\ Class Files,~Vol.~14, No.~8, August~2021}%
{Shell \MakeLowercase{\textit{et al.}}: A Sample Article Using IEEEtran.cls for IEEE Journals}


\maketitle

\begin{abstract}
A significant challenge facing current optical flow and stereo methods is the difficulty in generalizing them well to the real world. This is mainly due to the high costs required to produce datasets, and the limitations of existing self-supervised methods on fuzzy results and complex model training problems. To address the above challenges, we propose a unified self-supervised generalization framework for optical flow and stereo tasks: Self-Assessed Generation (SAG).
Unlike previous self-supervised methods, SAG is data-driven, using advanced reconstruction techniques to construct a reconstruction field from RGB images and generate datasets based on it. Afterward, we quantified the confidence level of the generated results from multiple perspectives, such as reconstruction field distribution, geometric consistency, and structural similarity, to eliminate inevitable defects in the generation process. We also designed a 3D flight foreground automatic rendering pipeline in SAG to encourage the network to learn occlusion and motion foreground.
Experimentally, because SAG does not involve changes to methods or loss functions, it can directly self-supervised train the state-of-the-art deep networks, greatly improving the generalization performance of self-supervised methods on current mainstream optical flow and stereo-matching datasets. Compared to previous training modes, SAG is more generalized, cost-effective, and accurate.          
\end{abstract}


\section{Introduction}
\IEEEPARstart{U}{nderstanding} the structure and motion of 3D scenes from a pair of images has been a long-standing goal of computer vision\cite{hartley2003multiple,sun2010secrets}. It is an essential cornerstone of many real-world applications, such as dynamic scene reconstruction technology\cite{10550869,wu20244d}, human motion recognition and prediction\cite{lertniphonphan2011human,arshad2022human,fan2018end}, virtual reality technology, and autonomous driving\cite{capito2020optical,wang2021end,wang2021learning}.

Traditional methods\cite{beauchemin1995computation,hirschmuller2007stereo} usually treat these tasks as energy minimization problems, such as optical flow variation methods and stereo semi-global optimization. Although traditional methods have achieved significant results, they cannot cope with occlusions, thin structures, and textureless areas. In recent years, deep learning methods\cite{sun2018pwc,teed2020raft} have become mainstream solutions for optical flow and stereo-matching tasks, achieving performance far superior to traditional methods in multiple benchmark tests\cite{Menze2015ISA,butler2012naturalistic}.


Compared with traditional methods, deep methods can learn robust matching and contextual features from massive labeled data to solve occlusion\cite{jiang2021learning} and textless problems better. However, building real-world datasets takes much work. Additional sensors such as LiDAR, RTK, IMU, and a sizeable manual annotation are required\cite{Menze2015ISA}. This leads to a very limited number of existing real-world datasets, limiting the application of deep methods in the real world\cite{xu2022gmflow}. Although recent research on synthetic datasets and self-supervised learning has, to some extent, alleviated the problem of data scarcity, they are still subject to constraints in domain transferability\cite{mayer2016large,butler2012naturalistic,mehl2023spring}, non-Lambertian surfaces, and indirect losses\cite{jonschkowski2020matters,kong2022mdflow}.

\begin{figure}[!t]
	\centering
	\includegraphics[width=3.4in]{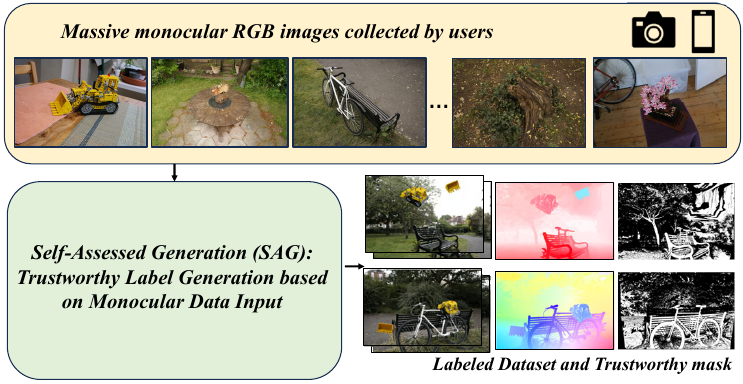}
	\caption{\textbf{The Idea of Data-driven Self-supervised Training.} We integrate optical flow and stereo matching into a unified framework (SAG) for self-supervised training. The core feature of SAG is data-driven, which extracts 3D structures from readily available monocular camera images or videos, filters out abnormal parts,  and generates customized datasets for 3D tasks. Unlike previous methods, SAG does not involve self-supervised loss and directly trains the model using generated data, achieving generalization performance and generality beyond loss-driven self-supervised schemes.}
	\label{fig_0}
\end{figure}
In this era where data is the gold mine, a massive and reliable dataset is a necessary factor for the success of powerful models such as GPT\cite{brown2020language} and SAM\cite{kirillov2023segment}. However, in optical flow and stereo matching, how to quickly and low-cost mine high-quality training data from the "gold mine" to fully tap into the enormous potential of deep networks is still a challenge that needs to be explored\cite{zhang2022clip}. In this paper, we propose a novel self-supervised training method, Self-Assessed Generation (SAG), to address this issue. As shown in Fig.\ref{fig_0}, SAG only uses photos or videos captured by a single camera as input, without manual intervention, and can quickly produce datasets at low cost to train various optical flow and stereo deep networks. Specifically, SAG consists of the following key modules:

\textit{Labeled Data Generation:}
This step aims to generate a preliminary dataset from the collected scenes. Using advanced NeRF\cite{barron2023zip} or 3DGS\cite{Yu_2024_CVPR} reconstruction techniques, precise 3D scenes are reconstructed from input image sets, and corresponding optical flow/stereo labels for image pairs are calculated from the camera's controllable motion.

\textit{Generate Data Self-assessment:}
In the process of scene reconstruction, inaccurate reconstruction and failure are inevitable, which further leads to incorrect generation results. Therefore, it is crucial to build a mechanism for assessing generated data. Before us, the only index used was ambient occlusion (AO)\cite{truong2023sparf,tosi2023nerf}. However, AO only applies to floating object detection in a few NeRF methods and cannot adapt to the rapidly developing reconstruction technology. Starting from the essence of the rendering formula, we propose the first universal reconstruction field quality evaluation index RC, which describes the degree of confidence of the reconstruction model in the reconstruction results from a probability perspective. Experimental results in Tab.\ref{tab:addlabel} and Tab.\ref{tab:stereoab} have shown that compared with other common indicators, such as geometric consistency and visual structure similarity, RC significantly improves model performance.

\textit{3D Flight Foreground Rendering Pipeline:}
In the previous steps, the generated optical flow results were caused by camera motion in a static scene, lacking a realistic motion foreground. Therefore, we propose a novel 3D foreground automatic generation pipeline. Based on SAM, a bidirectional confidence matching matrix is constructed to extract independent objects and their corresponding training labels from image pairs. The extracted independent moving objects will be randomly added to the initial image pair, actively creating occlusion while encouraging the optical flow network to learn more challenging motion prospects.
In addition, the 3D foreground generation pipeline is also suitable for stereo-matching. In stereo-matching, we generate foreground pairs that are clearly layered with the background, overlaying them on top of the initial stereo image pairs to provide stereo pairs with richer foreground layers. We have proven in experiments that in addition to effectively improving the network's reasoning ability for occlusion and foreground areas, the 3D flying foreground can also be applied to existing synthetic datasets to directly and effectively improve the generalization effect.

The final generated labeled data can be directly used for training deep networks\cite{sun2021loftr,xu2022gmflow,liu2022camliflow,ling2023learning,xu2023iterative,xu2023accurate,xu2023unifying}. As shown in Fig.\ref{fig_01}, the optical flow and stereo models trained by our SAG framework have achieved amazing zero-shot generalization effects. We believe that SAG is an essential step in driving optical flow and stereo models toward real-world applications, and it is also a stepping stone for establishing large matching models. We have summarized the characteristics of the SAG framework as follows:
\begin{figure}[!t]
	\centering
	\includegraphics[width=3.4in]{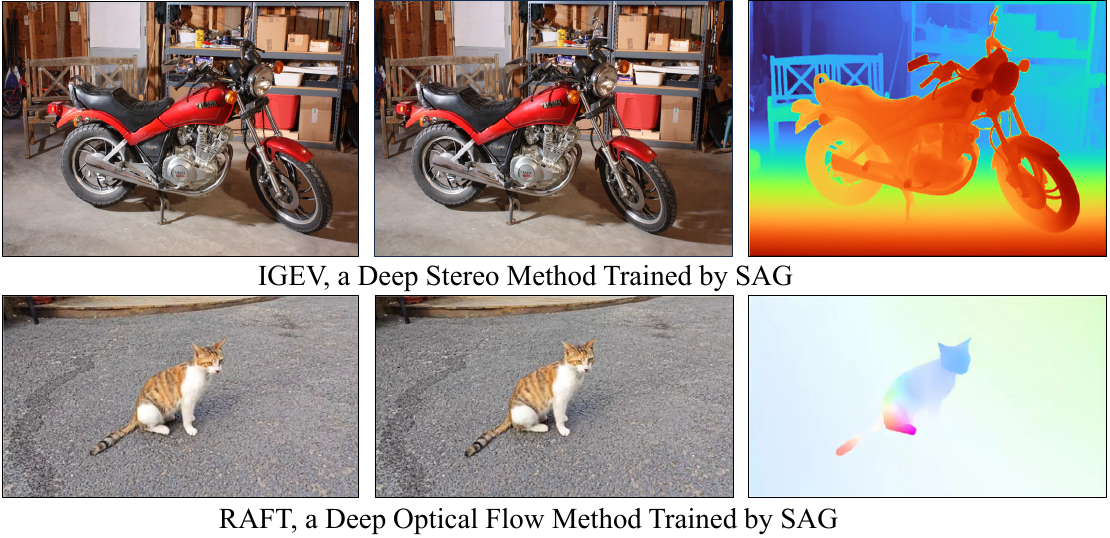}
	\caption{\textbf{Zero-shot Generalization of SAG in the Real World.} SAG achieved stunning real-world zero-shot generalization effects using only user-collected monocular RGB images. Moreover, unlike previous loss-driven self-supervised methods, SAG suits most existing methods.}
	\label{fig_01}
\end{figure}
\begin{enumerate}[1)]
	\item \textit{Compatibility:} SAG is compatible with almost all reconstruction methods based on NeRF and 3DGS frameworks, and can continue to benefit from the development of reconstruction methods.
	\item \textit{Low Cost:} Users only need to take photos or videos in the target application scenario, and corresponding datasets can be generated with just one click without any manual intervention.
	\item \textit{Data-driven Self-supervision: } SAG generates labeled data from input RGB image sets for training, making it applicable to almost all depth models, while previously self-supervised models\cite{kong2022mdflow,jonschkowski2020matters,hur2021self,luo2021upflow} based on indirect loss often could only train customized lightweight networks. Moreover, we have also demonstrated through experiments that with the continuous growth of data volume, the comprehensive generalization performance of SAG will continue to improve.
	\item \textit{High Generalization Precision:} We only used about 300 collected natural scenes to generate the training set, and the final trained model has achieved excellent generalization performance. Specifically, SAG has created a new first-class level of generalization in the real-world optical flow and stereo-matching dataset KITTI. On Sintel's public test platform, SAG reduced optical flow endpoint errors by 48\% (2.04 v.s. 3.96) compared to previous state-of-the-art methods, and even outperformed some supervised fine-tuning methods.
\end{enumerate}

This work is a substantial extension of our previous paper, ADFactory\cite{Ling_2024_CVPR}, presented at the CVPR2024 conference. The initial work presents an optical flow self-supervision scheme based on NeRF, which replaced traditional loss-driven methods.
This work proposes a more general and robust perspective that unifies optical flow and stereo-matching tasks into a data-driven, self-supervised training framework. The new framework is compatible with NeRF and 3DGS methods, addressing the issue of ADFactory needing real-world prospects. The new contributions are summarized below:
(1) To address the issues of slow NeRF baseline generation speed and poor geometric consistency of generated data in ADFactory, we introduced advanced 3DGS baselines and reconstructed the overall framework. The new framework significantly improves generation efficiency and data quality thanks to the explicit representation of the 3DGS baseline and rasterization rendering.
(2) We propose a 3D foreground automatic generation pipeline to address the lack of real foreground in ADFactory. Encourage the network to learn more challenging foregrounds while enhancing the learning of occlusion.
(3) In response to the potential generalization requirements of stereo-matching, we have extended ADFactory to stereo-matching tasks and conducted extensive experiments.
(4) We conducted a detailed ablation study on the new framework, particularly examining the advantages and disadvantages of different reconstruction baselines and depth rendering methods.
(5) We further demonstrated the superiority of our approach in the field of optical flow on the Sintel dataset. 
(6) We propose a simple and practical image quality assessment method EsFFT that can effectively remove low-quality images from the input image set.
Our code and video demo are available at \url{https://github.com/HanLingsgjk/UnifiedGeneralization}.

\begin{figure*}[!t]
	\centering
	\includegraphics[width=7in]{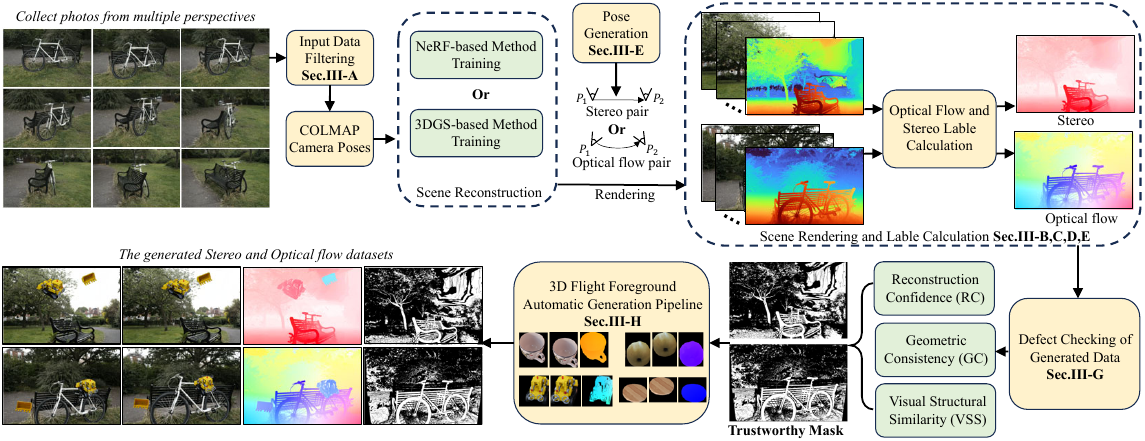}
	\caption{\textbf{SAG Pipeline.} Firstly, based on user-collected images, 3D scenes are reconstructed using 3DGS/NeRF and rendered to obtain RGB image pairs, depth maps, reconstruction confidence (RC), and occlusion. Afterward, the label calculation module will calculate the corresponding task label based on the previous rendering result and input the calculated label into the defect detection module to remove the defective part. The final generated label will also cover the 3D flight foreground to compensate for the insufficient foreground in the generated dataset.}
	\label{fig_method}
\end{figure*}

\section{Research background}
In this section, we introduce some research closely related to SAG to support the motivation of this article.
\subsection{Synthetic Dataset}
The synthetic dataset\cite{mayer2016large,butler2012naturalistic} is extracted from manually constructed 3D animation scenes and is the essential training method for existing depth methods. However, data domain bias is inevitable when testing in natural scenes. Some companies have invested enormous amounts of workforce and GPU computing power to bridge these data domain biases and build synthetic scenarios as realistic as possible for network training, achieving certain results, but these valuable data assets are rarely open source and cannot be widely used.

The SAG proposed in this paper can quickly build a fine-tuning dataset for the target domain at a low cost to alleviate the inherent data bias and high cost of the synthesized data mentioned above.
\subsection{Loss-driven Self-supervision Methods}
The previous self-supervised learning methods\cite{kong2022mdflow,jonschkowski2020matters,hur2021self,luo2021upflow} relied on photometric loss and structural similarity loss (SSIM), which can be trained on any pair of images collected by users. It is one of the feasible paths for deep methods to generalize in the real world. However, many key issues remain to be solved: (1) Due to indirect loss training, the results obtained by self-supervised methods are generally vague and unclear.
(2) Self-supervised networks are often specially designed lightweight networks. Also, because of indirect losses, it is difficult for existing self-supervised methods to train complex networks. (3) Self-supervised loss cannot handle ill-posed problems such as occlusion and non-Lambertian surfaces.
On the contrary, SAG trains deep networks based on data-driven self-supervision, which can adapt to almost any model. With the development of reconstruction technology\cite{verbin2022ref,attal2021torf,luiten2023dynamic,yang2023emernerf,wu2022d}, there is hope to alleviate ill-posed problems (reflective surfaces, transparent surfaces).

\subsection{Generation method based on 3DGS/NeRF}
There have been some reconstruction-based dataset generation works before us. Yang et al.\cite{yang2023unisim,yang2023emernerf} used vehicle-mounted cameras and radars as data input and performed scene reconstruction based on NeRF. Combining the background field, foreground field (cars and pedestrians), and sports field (scene flow/optical flow) obtained by self-supervised training to render RGB images dynamic scenes and corresponding depth/optical flow labels.
However, they mainly focus on the appearance of RGB rendering results, lacking an effective result filtering mechanism. In addition, additional sensors (Lidar) further increase the cost of data production.
Tosi et al.\cite{tosi2023nerf} also proposed a semi-data-driven stereo network training method, NS-stereo, which directly trains the stereo network by rendering and calculating new stereo pairs and labels from the static scenes reconstructed by NeRF, achieving good generalization results on the Middlebury\cite{scharstein2014high} dataset. The main problem with NS-stereo is that self-supervised loss was used during training without achieving true data-driven, and comprehensive data quality evaluation metrics were not proposed. Moreover, the special training and loss composition make it inconvenient for NS to promote quickly. In comparison, SAG is the first pure data-driven self-supervised training framework that is compatible with NeRF and 3DGS. A single reconstruction can produce datasets suitable for multiple tasks and comprehensively evaluate the quality of the generated datasets.

\section{Method}
In this section, we will introduce the overall process of SAG. As shown in Fig.\ref{fig_method}, our overall method mainly consists of five parts: Preprocessing the input data to remove low-quality images (Sec.\ref{IDF}); Reconstruct and representation of a 3D scene from input images using 3DGS or NeRF (Sec.\ref{sec:gs} and Sec.\ref{sec:nerf}); Generate camera poses based on different tasks, rendering and calculating labeled data from the reconstructed scene according to the poses (Sec.\ref{sec:md}, Sec.\ref{seclable} and Sec.\ref{sec:occ}); Perform quality checks on the data and remove any reconstructed areas not up to standard (Sec.\ref{mask_check}); Based on the 3D foreground automatic rendering pipeline, the generated data is overlaid with flying foreground to compensate for the problem of insufficient foreground (Sec.\ref{3Dfore}).

Please note that flight foreground generation is an independent process. We generate a 3D flight foreground database in advance during the experiment and then overlay it on the generated results. In Sec.\ref{sec:gs} and Sec.\ref{sec:nerf}, we mainly introduced the representation and rendering methods of 3DGS and NeRF. For more information on scene reconstruction, please refer to \cite{barron2023zip} and \cite{kerbl20233d}.
\subsection{Input Data Filtering} \label{IDF}
Clear and sharp photo input is necessary for COLMAP\cite{schoenberger2016sfm,schoenberger2016mvs}, NeRF, and 3DGS. However, while collecting data, we found that videos captured casually inevitably exhibit motion blur, which poses significant challenges to scene reconstruction. In previous works, this topic has been rarely mentioned. Although there are some reconstruction methods\cite{Ma_2022_CVPR} specifically for blurred scenes, we hope that the scene reconstruction part in the framework can be replaced at any time to enjoy the dividends brought by the rapid development of NeRF and 3DGS methods. 

To address this difficulty, we propose an improved version of the Fast Fourier Transform (FFT) fuzzy detection scheme, edge selection FFT (EsFFT). The fuzzy detection method based on FFT is a ubiquitous and straightforward blur detection algorithm. Its core idea is to analyze the overall high-frequency content in the image through FFT and set a threshold for the high-frequency content to determine whether the image is blurry. The biggest problem with its use is that it is difficult to set a universal threshold for all scenes. In some cases, the high-frequency components of motion-blurred images with rich textures will be much higher than those of clear images with sparse textures. Our proposed solution only considers the high-frequency content near the textures, significantly improving the FFT scheme's generalization performance. We have published source code\footnote{\url{https://github.com/HanLingsgjk/EsFFT}} for readers to test. 
The use of EsFFT is for the sake of lightweight and practicality. In our actual use process, it is already sufficient to handle most daily scenarios. If there is further demand, more advanced depth methods\cite{tang2019defusionnet} can be used.

\subsection{Scene Representation and Rendering of NeRF}\label{sec:nerf}
In this section, we introduce the scene representation and rendering process of NeRF class methods based on the commonly used Zip-NeRF\cite{barron2023zip}.
 Zip-NeRF calculates the color and depth of specific pixels through volume rendering based on ray sampling. Specifically, the complete scene is usually stored in a hash voxel field in INGP\cite{muller2022instant}, which maps the interval distance $T_i = [t_i,t_{i+1})$ on the ray $\bm{\mathrm{r}}(t) = \bm{\mathrm{o}} + t\bm{\mathrm{d}}$ into a set of color features $\bm{\mathrm{c}} \in [0,1]^3 $ and volume density $\sigma \in \mathbb{R}^+ $, where $\bm{\mathrm{o}}$ and $\bm{\mathrm{d}}$ are the origin and direction of the ray, respectively, $t$ is the distance from the origin along the ray direction. It can be formulated as:
\begin{equation}
	(\sigma_i,\bm{\mathrm{c}}_i) = \mathrm{MLP}_\theta(\gamma(\bm{\mathrm{r}}(T_i))) , \quad\forall T_i \in \bm{\mathrm{t}}
\end{equation}
where, $\gamma$ is a mapping sampling module that maps the ray segment $\bm{\mathrm{r}}(T_i)$ to a conical surface and samples the corresponding position features in INGP, $\mathrm{MLP}_\theta$ is a shallow MLP with parameter $\theta$ responsible for mapping the features sampled by $\gamma$ to $(\sigma_i,\bm{\mathrm{c}}_i)$, and $\bm{\mathrm{t}}$ is the set of all intervals on the ray that are included in the rendering.

With these volume densities and color features, we can calculate the pixel color $\bm{\mathrm{C}}$ corresponding to the ray based on the volume rendering formula:
\begin{equation}
	\bm{\mathrm{C(r,t)}} = \sum_i w_i \bm{\mathrm{c}}_i
	\label{eq:C1}
\end{equation}

\begin{equation}
	w_i=\left(1-e^{-\sigma_i\left(t_{i+1}-t_i\right)}\right) {E}_{i}
	\label{eq:C2}
\end{equation}

\begin{equation}
	E_{i} = e^{-\sum_{j<i} \sigma_{j}\left(t_{j+1}-t_{j}\right)}
\end{equation}
where $\left(1-e^{-\sigma_i\left(t_{i+1}-t_i\right)}\right)$ is the opacity at $t_i$, $E_{i}$ is the cumulative transmittance along the ray starting point to the $t_i$ position. According to the construction of Eq.\ref{eq:C1}, the sum of the weights $w_i$ on a ray is always less than or equal to 1\cite{barron2022mip}. When the ray points towards an opaque surface, the sum of the weights $w_i$ approaches 1. \textbf{Therefore, Eq.\ref{eq:C2} can be seen as calculating the expected color of a ray.}

The depth calculation formula corresponding to the ray can be written as:
\begin{equation}
	z\bm{\mathrm{(r,t)}} = \sum_i w_i t_{mid}, \quad t_{mid}=\frac{(t_i+t_{i+1})}{2} 
	\label{eq:eq5}
\end{equation}   
here, $z\bm{\mathrm{(r,t)}}$ is the mean depth (expected depth) of ray $\bm{\mathrm{r}}$, and the meaning of $w_i$ is the probability of the existence of a surface at position $t_i$ on the ray.

\subsection{Scene Representation and Rendering of 3DGS}\label{sec:gs}
Unlike NeRF-like methods, existing 3DGS methods\cite{Yu_2024_CVPR,kerbl20233d} express the entire scene through a set of explicit 3D gaussian primitives $\mathcal{G}=  \{  g_1,g_2,...,g_N \}  $ and complete the rendering of the entire image at once using volume splatting technology. The geometric properties of gaussian primitive $g_k$ in space can be parameterized as:
\begin{equation}
	g_k(\bm{x})=e^{-\frac{1}{2}\left(\bm{x}-\mathbf{p}_k\right)^T \boldsymbol{\Sigma}_k^{-1}\left(\bm{x}-\mathbf{p}_k\right)}
	\label{eq:gs}
\end{equation}
where $\mathbf{p}_k \in \mathbb{R}^{3 \times 1}$ is the center position of $g_k$, and $\boldsymbol{\Sigma}_k\in \mathbb{R}^{3 \times 3}$ is the semi-definite covariance matrix that constrains the rotation direction and scale of the gaussian body. In addition, each gaussian primitive includes a color parameter $CF_k$ based on the spherical harmonic function and an opacity parameter $\alpha_k \in[0,1]$.

When rendering an image, the first step is to project gaussian primitives onto the corresponding imaging plane based on the camera's internal and external parameters. The projected gaussian primitives can be referred to as $g_k^{2D}(\bm{x})$:

\begin{equation}
	g_k^{2D}(\bm{x})=e^{-\frac{1}{2}\left(\bm{x}-\mathbf{p}_k\right)^T  \left(\boldsymbol{\Sigma}_k^{2 D}\right)  ^{-1}\left(\bm{x}-\mathbf{p}_k\right)}
\end{equation}
where $\boldsymbol{\Sigma}_k^{2 D}$ is a 2D covariance matrix calculated based on camera pose and $\boldsymbol{\Sigma}_k$, See 3DGS\cite{kerbl20233d} for details.

Finally, 3DGS calculates the color of pixels by blending $g_k^{2D}(\bm{x})$ sorted by depth ($1 \rightarrow K$):
\begin{equation}
	\bm{\mathbf{C}(x)}=\sum_{k=1}^K w_k\mathbf{c}_k 
\end{equation}
\begin{equation}
	w_k= \alpha_k g_k^{2D}(\bm{x}) \prod_{j=1}^{k-1}\left(1-\alpha_j g_j^{2D}(\bm{x})\right)
\end{equation}
among them, $\mathbf{c}_k $ is calculated based on spherical harmonic parameters $CF_k$ and corresponding viewing angles.

Similar to NeRF, the mean depth rendering of 3DGS can be obtained from the following equation:
\begin{equation}
	z(\bm{x})=\sum_{k=1}^K w_k \bm{d}_k 
	\label{eq:eq10}
\end{equation}
where $\bm{d}_k$ is the distance from the center of the gaussian primitive $g_k$ to the projection plane.

\begin{figure}[!t]
	\centering
	\includegraphics[width=3.4in]{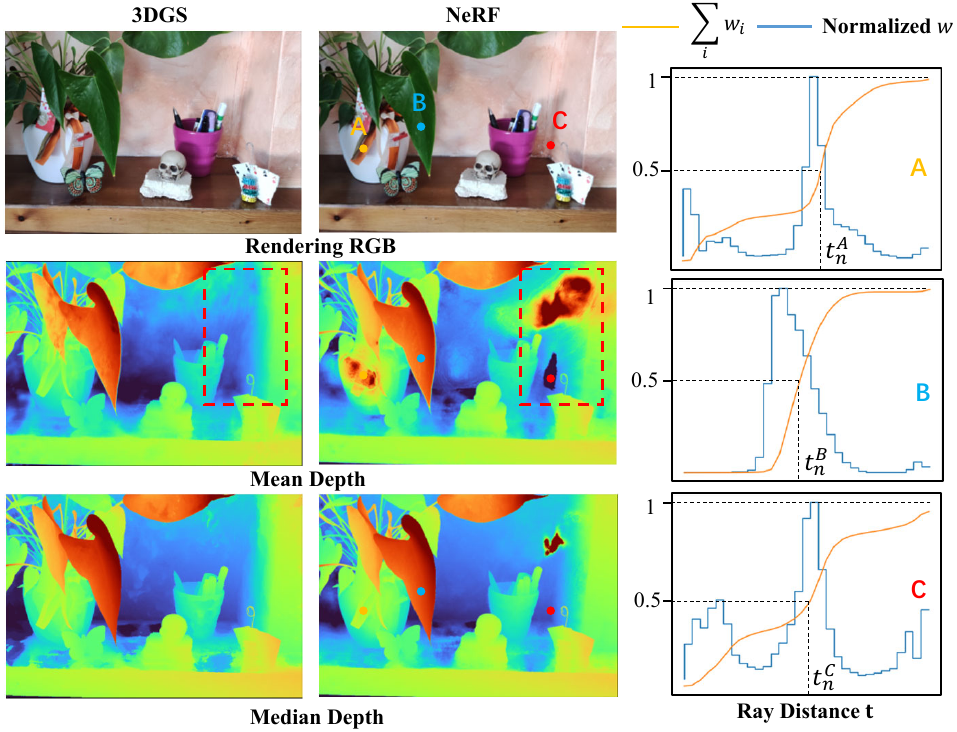}
	\caption{\textbf{Median Depth v.s. Mean Depth.} Left: Rendered RGB image and two different depths of 3DGS and NeRF. Right: The weights $w$ on three different rays (based on NeRF): well-trained ray B, ray C with incorrect surfaces, and ray A with potential multiple surfaces.
		We found that the mean depth in A and C is clearly incorrect, as the weighted depth of the wrong surface interfered with the final result, especially when there were incorrect weights at the far end of the ray. 
		And the median depth can reduce the interference of these erroneous surfaces.}
	\label{fig_1}
\end{figure}
\subsection{Median Depth} \label{sec:md}
Although most generative methods\cite{tosi2023nerf,truong2023sparf,yang2023unisim,yang2023emernerf} based on NeRF or 3DGS use the depth described in Eq.\ref{eq:eq5} or Eq.\ref{eq:eq10}, this depth is unreasonable. As mentioned in Sec.\ref{sec:nerf}, its definition is the expected distance between all potential surfaces on a ray. Therefore, when there are multiple surfaces or insufficient reconstruction areas on a ray, it often fails.

In this article, we use median depth to calculate the labels of optical flow and stereo, which are defined as follows:
\begin{equation}
	z\bm{\mathrm{(r,t)}} = t_n, \quad   \sum_{i=1}^n w_i = 0.5
	\label{eq:dnerf}
\end{equation} 
\begin{equation}
	z(\bm{x})=\bm{d}_m, \quad    m=\arg\min_{l \in \mathbb{N}} \left| \sum_{k=1}^l w_k - 0.5\right|  
	\label{eq:dgs}
\end{equation}

Eq.\ref{eq:dnerf} and Eq.\ref{eq:dgs} are the median depth calculation formulas for NeRF and 3DGS, respectively. The meaning of Eq.\ref{eq:dnerf} is to take the distance $t_n$ as the depth when the cumulative value of the weight $w$ is 0.5. NeRF is continuously sampled in practical calculations to calculate the median position through interpolation. However, unlike NeRF, 3DGS is very discrete, so we directly take the distance of the gaussian primitive closest to the median as the depth.

As shown in Fig.\ref{fig_1} B, the sum of weight $w$ on a well-trained ray should appear as a step function. At this point, the median depth position is close to the peak of $w$ (Location with the highest potential surface probability). As shown in Fig.\ref{fig_1} A and C, when multiple surfaces and incorrect surfaces are on the ray, the median depth can better avoid errors and approach the correct depth.

Furthermore, we conducted quantitative ablation experiments in  Sec.\ref{mmu} to verify the rationality of using median depth. In the experiment, we found that compared to NeRF-based methods, the 3DGS method achieved better geometric consistency thanks to its compact point representation. The compact point representation also makes the depth obtained by 3DGS rendering less prone to artifacts, as shown in the red box in Fig.\ref{fig_1}.

\subsection{Labeled Data Generation for Optical Flow and Stereo}\label{seclable}

The complete labeled optical flow data consists of two consecutive RGB frames and an optical flow field, which reveals the pixel coordinate changes of points in space between the two frames. The labeled stereo data consists of two parallel RGB images and their corresponding disparity fields. Specifically, we first generate corresponding random camera poses for the task, then render the corresponding RGB image and depth in the trained scene based on the poses, and finally calculate the optical flow field and disparity field based on the rendering results.

\textbf{Camera Pose Generation:} The camera poses $\bm{P}$ is a $4 \times 4$ matrix composed of a rotation matrix $\bm{R}$ and a  position matrix $\bm{T}$, where $\bm{R}$ and $\bm{T}$ constrain the camera's orientation and position. Specifically, we recommend not directly generating a random camera pose, as each scene's scale and voxel distribution are different, and random poses are likely to be outside the scene. The new camera pose $\bm{P}_2$ in our paper comes from fine-tuning the orientation and position based on the existing pose  $\bm{P}_1$.
\begin{equation}
	\bm{P}_2 = \bm{P}_1\bm{R}_r+\bm{T}_r
\end{equation}
where $\bm{R}_r$ and $\bm{T}_r$ are the fine-tuning matrices for orientation and position, respectively. For the optical flow task, $\bm{R}_r$ and $\bm{T}_r$ will be randomly assigned within a small amplitude. For the stereo task, $\bm{R}_r$ is the identity matrix, and the $\bm{T}_r$ matrix only has one fine-tuning $\Delta d$ in the horizontal position to form a stereo pair.

\textbf{Optical Flow Label Calculation:}
After getting the camera pose pair $\bm{P}_1$ and $\bm{P}_2$, we can render the corresponding RGB images $I_1, I_2$  and depth $Z_1, Z_2$ based on the NeRF or 3DGS techniques described in \ref{sec:nerf}, \ref{sec:gs} and \ref{sec:md}. Taking NeRF as an example, where $I_1(u,v) = \bm{\mathrm{C(r,t)}}$, $Z_1(u,v) = z\bm{\mathrm{(r,t)}}$, $(u,v)$ is the pixel plane coordinate corresponding to ray $\bm{\mathrm{(r,t)}}$.

\begin{figure}[!t]
	\centering
	\includegraphics[width=2.6in]{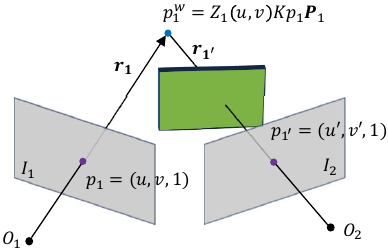}
	\caption{\textbf{Optical Flow Calculation and Occlusion.} The optical flow calculation is divided into two steps. Firstly, point $p_1$ in the pixel plane of the first frame is projected onto point $p_1^{w}$ in the world coordinate system. Then, based on the pose $\bm{P}_2$, the point $p_1^{w}$ is projected onto the pixel plane of the second frame to obtain point $p_{1'}$. This figure also shows the occlusion situation, with a solid surface between $p_1^{w}$ and the camera's optical center $O_2$. At this time, we can evaluate whether ray $\bm{r_{1'}}$ is occluded by calculating the integral value of the weight $w$ between $p_1^{w}$  and  $O_2$.}
	\label{fig_2}
\end{figure}

As shown in Fig.\ref{fig_2}, based on the $Z_1$ and the camera poses $\bm{P}_1,\bm{P}_2$, we can calculate the position of the pixel $p_1=(u,v,1)$ in the first frame to corresponding $p_{1'}=(u',v',1)$ in the second frame, as follows:
\begin{equation}
	\begin{aligned}
		   Z_{1'}(u,v)Kp_{1'}\bm{P}_2 &=Z_1(u,v)K p_1 \bm{P}_1  \\
		  Z_{1'}(u,v) p_{1'}&= K^{-1}Z_1(u,v)Kp_1\bm{P}_1 {\bm{P}_2}^{-1} \\
		  p_{1'} &= \frac{K^{-1}Z_1(u,v)Kp_1\bm{P}_1 {\bm{P}_2}^{-1}}{Z_{1'}(u,v)}
	\end{aligned}
	\label{eq16}
\end{equation}   
\begin{equation}
	f_{1\rightarrow 2}(u,v) = p_{1'} - p_{1}
	\label{eq:eq8}
\end{equation}  
where $ f_{1\rightarrow 2}$ is the optical flow between frames $1$ and $2$, $K$ is the camera intrinsic matrix. Please note that $Z_{1'}(u,v)$ in Eq.\ref{eq16} is a real number calculated by $K^{-1}Z_1(u,v)Kp_1\bm{P}_1 {\bm{P}_2}^{-1}$, and due to the special construction of $p_{1'}$, it can be divided into the denominator.

\textbf{Stereo Label Calculation:} After obtaining a pair of horizontal poses $\bm{P}_1,\bm{P}_r$, we can refer to Eq.\ref{eq16} to calculate the optical flow between two frames. The absolute value of the optical flow is equivalent to the disparity $d$, which is:
\begin{equation}
	d = \left|f_{l\rightarrow r}\right|
	\label{eq:eqstereo}
\end{equation}  

There is another simpler method for calculating disparity:

\begin{equation}
	d = \frac{b \cdot{f}}{Z_1}
	\label{eq:eqstereo2}
\end{equation}  
where $b$ is the camera baseline, and $f$ is the camera focal length comes from the COLMAP.

\subsection{Calculation of Occlusion Mask} \label{sec:occ}
Detecting and handling occlusion has always been one of the core difficulties in optical flow and stereo tasks, especially in the research of self-supervised methods\cite{luo2021upflow,kong2022mdflow}, where accurate occlusion detection is a key factor in determining method performance. This section will explain how to calculate occlusion masks based on NeRF and 3DGS, respectively.

\textbf{NeRF:}
As shown in Fig.\ref{fig_2}, in this paper, we determine whether there is occlusion between two frames by calculating the cumulative probability of surfaces on the new ray $\bm{r_{1'}}$:
\begin{equation}
	OCC(u,v) = \sum_{j=1}^{n-1} w_j, \quad  n=\arg\min_{l \in \mathbb{N}} \left| t_l - Z_{1'}(u,v) \right| 
	\label{eq:eqocc}
\end{equation}  

\begin{equation}
	M_{occ}(u,v)= \begin{cases}0 & \text { if } OCC(u,v)>th \\ 1 & \text { otherwise }\end{cases}
\end{equation}
where $OCC(u,v)$ is the probability of a surface between the viewpoint $O_2$ and the  $p_1^{w}$. $M_{occ}$ is the occlusion mask between two frames (when the value is 0, it indicates the presence of occlusion), and in experiments, $th$ is generally set to 0.3.

\textbf{3DGS:}
Due to the inherent method characteristics of 3DGS, it is not easy to calculate occlusion through weight integration on rays like NeRF. Therefore, we adopted the classic positive forward-backward consistency verification scheme. For more information, please refer to \cite{wang2018occlusion}.

\subsection{Defect Checking of Generated Data} \label{mask_check}
In this section, we evaluate the quality of the generated dataset from multiple perspectives.  It is worth mentioning that we innovatively propose a quality evaluation index that is almost applicable to all NeRF or 3DGS reconstruction methods: Reconstruction confidence (RC). In the following part, we first introduce the essential RC indicator (the module that significantly improves performance in ablation experiments),  then sequentially introduce the remaining geometric consistency (GC) and visual structural similarity (VSS) indicators.

\begin{figure}[!t]
	\centering
	\includegraphics[width=3.4in]{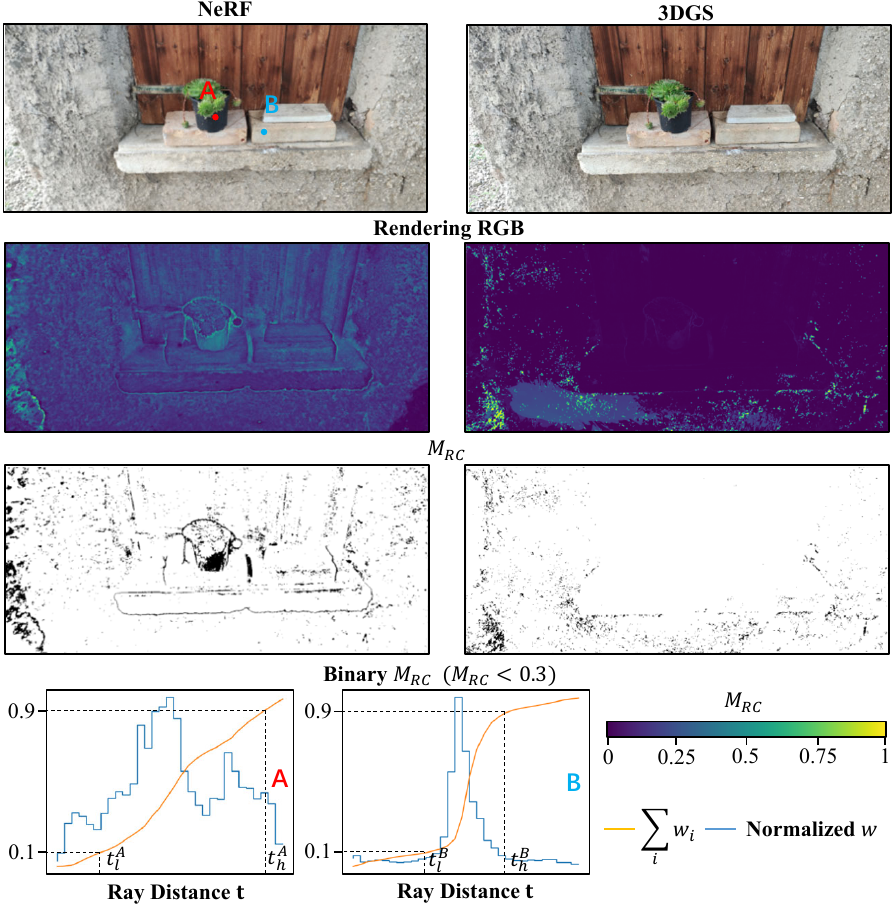}
	\caption{\textbf{Visualization of Reconstruction Confidence (RC).} Top: RGB image rendered by NeRF and 3DGS, $M_{RC}$ confidence mask, and binarized $M_{RC}$. Below: A represents the weight distribution of rays in dark areas without texture, and B represents the weight distribution of rays in rich texture areas. The $M_{RC}$ we proposed easily segmented those areas where the radiation field is difficult to reconstruct based on the ray weight distribution. In addition, it can be found that the overall aggregation degree of 3DGS is better than NeRF, thanks to the compact scene expression of 3DGS itself.}
	\label{figconf}
\end{figure}
\begin{figure}[!t]
	\centering
	\includegraphics[width=3.2in]{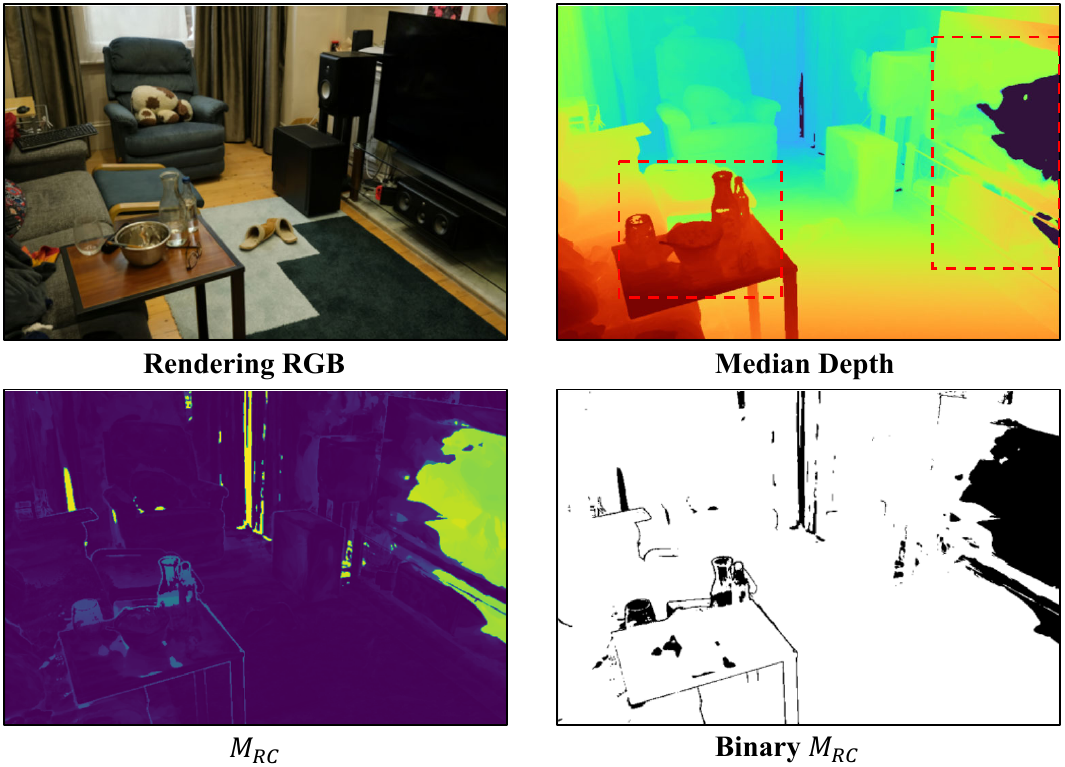}
	\caption{\textbf{RC Visualization of 3DGS Method.} We present a more challenging scene here, observing the red box in the median depth, which includes transparent glass products, reflective surfaces, and large areas of pure black. The depth map shows significant distortion in the reconstruction of these parts, and our proposed $M_{RC}$ accurately separates these unstable parts of the reconstruction.}
	\label{figgsconf}
\end{figure}

\textbf{Reconstruction Confidence (RC):} In 3DGS and NeRF generative application scenarios, the quality of scene reconstruction directly determines the method's overall performance. However, currently, yet to be a reliable evaluation indicator applies to most methods. To overcome this problem, we propose a widely applicable new indicator for measuring the quality of scene reconstruction: Reconstruction Confidence.

The original inspiration came from the observation of neural radiation fields (NeRF). As shown in Fig.\ref{figconf} B, when the neural field is well trained, the weight $w$ distribution of rays should be clustered at a central point, and the weight integration should appear as a step function, indicating that the radiation field has an exact surface on that ray. Moreover, we observed that such rays often correspond to areas with rich textures and appropriate brightness. On the contrary, it is difficult for the radiation field to learn the correct surface position for areas that are too dark or textureless.

Specifically, the calculation formula for RC in NeRF can be written as:
\begin{equation}
	M_{RC}^{nerf}(u,v) = \frac{t_h - t_l}{t_h + t_l}
	\label{eq:eq12}
\end{equation}   
\begin{equation}
	\sum_{i=1}^l w_i = th_{low}, \quad \sum_{j=1}^h w_j = th_{high}
	\label{eq:eq13}
\end{equation}  
as shown in Fig.\ref{figconf} A and B, $th_{low}$ and $th_{high}$ are boundary thresholds, which are generally set to 0.1 and 0.9 in experiments. $t_l$ and $t_h$ are the depths at which the boundary threshold is reached.

Similarly, the calculation formula for RC in the 3DGS method can be written as:
\begin{equation}
	M_{RC}^{3dgs}(u,v) = \frac{t_h - t_l}{t_h + t_l}
	\label{eq:eqgsrad}
\end{equation}   
\begin{equation}
		 l=\arg\min_{h \in \mathbb{N}} \left| \sum_{i=1}^h w_i - th_{low}\right| 
	\label{eq:eqgsconf}
\end{equation}  
\begin{equation}
	h=\arg\min_{h \in \mathbb{N}} \left| \sum_{j=1}^h w_j - th_{high}\right|
	\label{eq:eqgsconf2}
\end{equation}  

Essentially, RC is how confident the NeRF/3DGS method is in rendering depth values. Specifically, in the depth rendering formulas Eq.\ref{eq:dnerf} and Eq.\ref{eq:dgs}, $w$ can be interpreted as the probability of a solid surface being present somewhere. RC describes the degree of cohesion in the probability $w$ of the interval $(t_l, t_h)$ near the median depth obtained from rendering. Therefore, the smaller the value of RC, the more confident the reconstruction method is in the rendering results. As shown in Fig.\ref{figconf} and Fig.\ref{figgsconf}, RC can accurately distinguish the failed reconstruction parts in both NeRF and 3DGS methods.
\begin{figure*}[!t]
	\centering
	\includegraphics[width=6.8in]{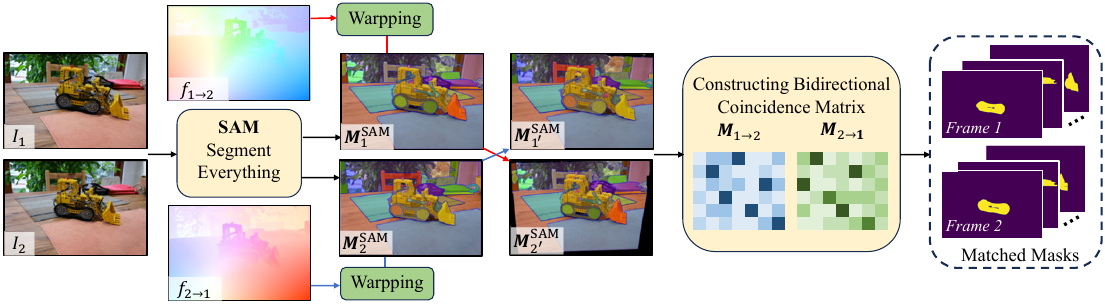}
	
	\caption{\textbf{3D Flight Foreground Automatic Generation Pipeline.} Based on the labels calculated from the rendering results, this module performs bidirectional matching and screening of the mask blocks generated by SAM to obtain matching 3D flight foregrounds.}
	\label{fig_3Dpipe}
\end{figure*}

\begin{figure}[!t]
	\centering
	\includegraphics[width=2.6in]{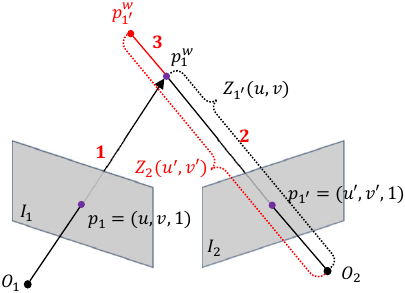}
	\caption{\textbf{Geometric Consistency Check.} We have shown the main three steps for calculating the GC value of point $p_1$. (1) Project point $p_1$ onto the world coordinate system to obtain $p_1^w$; (2) Project $p_1^w$ onto $I_2$ to obtain $p_{1'}$; (3) Combine $p_{1'}$ with the depth of $I_2$ and project it back into the world coordinate system to obtain $p_{1'}^w$. If the depth between two frames is completely consistent, there should be $p_{1'}^w=p_1^w$.}
	\label{figgc}
\end{figure}

\textbf{Geometric Consistency (GC):} GC describes the multi-view consistency of the rendering depth values obtained by the reconstruction method, which is one of the core indicators for stereo matching and optical flow label accuracy. Based on previous work\cite{schonberger2016pixelwise}, we have proposed deep consistency verification processes for NeRF and 3DGS methods, as shown in Fig.\ref{figgc}.

Next, taking point $p_1$ in Fig.\ref{figgc} as an example, we explain how to calculate geometric consistency. Firstly, based on the rendering depth $Z_1$ of $I_1$, we project $p_1$ onto the $p_1^w$ point in the world coordinate system:
\begin{equation}
	p_1^w = Z_1(u,v)K p_1 \bm{P}_1
	\label{eq:p1}
\end{equation}   

Then, project $p_1^w$ onto the pixel plane of $I_2$ to obtain the corresponding projection point $p_{1'}$:
\begin{equation}
	\begin{aligned}
	Z_{1'}(u,v) p_{1'} &= K^{-1} p_1^w {\bm{P}_2}^{-1} \\
	 p_{1'} &=\frac{K^{-1} p_1^w {\bm{P}_2}^{-1}}{Z_{1'}(u,v)} 
	\end{aligned}
	\label{eq:p2}
\end{equation}   

The third step is to sample the depth $Z_2(u',v')$ corresponding to $p_1'$ in $Z_2$ (in NeRF, $Z_2(u',v')$ can obtain more accurate results based on ray rendering of point $p_1'$), and project $p_1'$ back to the world coordinate system based on the sampled depth $Z_2(u',v')$:
\begin{equation}
	p_{1'}^w = Z_2(u',v')K p_{1'} \bm{P}_2
	\label{eq:p3}
\end{equation}   

If the geometric structures expressed by $I_1$ and $I_2$ are entirely consistent, then there should be $p_{1'}^w=p_1^w$. Since points $p_{1'}^w$, $p_{1}^w$ and $O_2$ are collinear, we calculate the geometric consistency as:
\begin{equation}
	M_{GC}(u,v) =\frac{\left| Z_{1'}(u,v) - Z_2(u',v')\right|}{ Z_2(u',v') + Z_{1'}(u,v)}
	\label{eq:p8}
\end{equation}  
where, the $M_{GC}$ value is between 0 and 1, and the closer it is to 0, the better the depth consistency.

\textbf{Visual Structural Similarity (VSS):}
This indicator directly checks the correctness of optical flow and stereo labels. Its basic principle is directly warpping $I_2$ into $I_1$ based on the current calculated transformation field (optical flow/stereo). If the transformation field is correct, the warpped $I_{1'}$ should be the same as $I_1$ (excluding occluded areas). In this article, we use SSIM\cite{
	wang2004image} to check the visual structural similarity between the warpped result and the original image:
\begin{equation}
	M_{VSS} =1- \mathrm{SSIM}(I_{1},I_{1'})
	\label{eq:ssim}
\end{equation}  
where $M_{VSS}$ is a structural similarity index with values between 0 and 1. The smaller the value, the higher the quality of optical flow matching. This indicator can also serve as a generalization performance evaluation indicator for optical flow/stereo, and needs to be combined with occlusion masks to remove artifacts when used.

\begin{figure}[!t]
	\centering
	\includegraphics[width=3.2in]{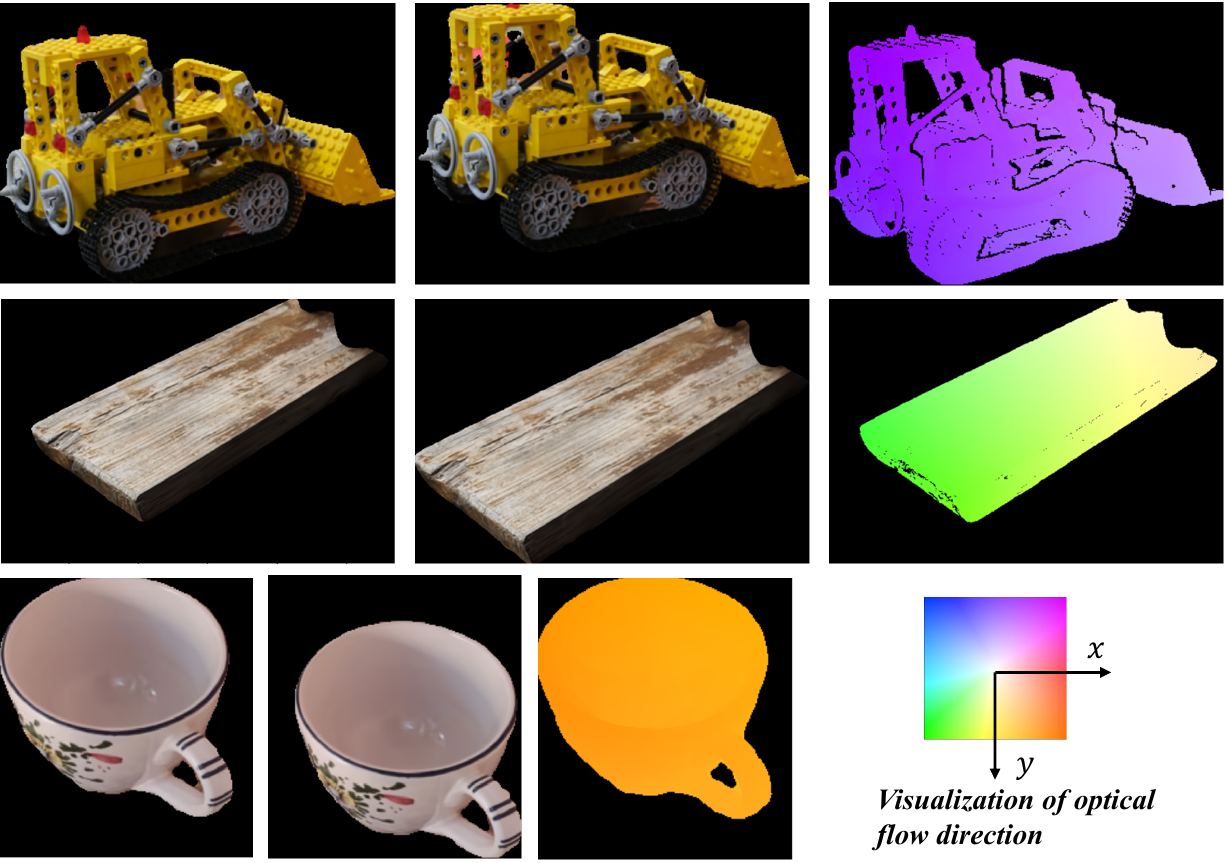}
	\caption{\textbf{
			Visualization of 3D Flight Foregrounds $F_{3d}$.} From left to right are the RGB image pairs of the flight foreground and the visualized optical flow labels (with errors removed).}
	\label{figfore3D}
\end{figure}

\subsection{3D Flight Foreground Automatic Rendering Pipeline} \label{3Dfore}

This section proposes a practical 3D foreground automatic generation scheme to compensate for the lack of real motion foreground in the generated data. The essence of this method is to accurately match object blocks between a pair of labeled image data, as shown in Fig.\ref{fig_3Dpipe}, we render a pair of RGB images based on the techniques proposed in the previous chapters, and calculate the forward and backward optical flow between these two frames, as well as the corresponding effectiveness masks. Then, the two frames of images are segmented into object blocks using SAM\cite{kirillov2023segment}, and bidirectional object block matching is performed by constructing a confidence matrix. After the matching is completed, we will further screen to obtain the final 3D foreground.

\textbf{Input:} Based on a reconstructed scene, we sequentially render and calculate a pair of RGB images $I_1$ and $I_2$, corresponding bidirectional optical flows $f_{1\rightarrow2}$ and $f_{2\rightarrow1}$, as well as a series of masks $M_{RC}$,$M_{GC}$ and $M_{VSS}$ as inputs. 

\textbf{Segment Anything Model (SAM):} We send $I_1$ and $I_2$ into SAM to obtain two sets of binarized segmentation masks $\bm{M}_{1}^{\mathrm{SAM}} \in \mathbb{R}^{N_{1} \times H \times W}$ and $\bm{M}_{2}^{\mathrm{SAM}} \in \mathbb{R}^{N_{2} \times H \times W}$ in everything mode, where $N$ is the number of binarized masks, which may not necessarily be equal in two consecutive frames (in Fig.\ref{fig_3Dpipe}, these masks are color-coded and overlaid on the RGB image).

\textbf{Mask Matching:} 
We aim to match masks belonging to the same object, but it is evident that the shape and position of these masks are not aligned between the two frames. Therefore, we use bidirectional optical flow $f_{1\rightarrow2}$, $f_{2\rightarrow1}$ to align  $\bm{M}_{1}^{\mathrm{SAM}}$, $\bm{M}_{2}^{\mathrm{SAM}}$ warp to corresponding $\bm{M}_{2'}^{\mathrm{SAM}}$ and $\bm{M}_{1'}^{\mathrm{SAM}}$, achieving bidirectional alignment of masks between two frames. Next, construct coincidence matrix to find the best matching mask:
\begin{equation}
	\bm{M}_{1 \rightarrow 2}(i,j) = {\frac{Sum(\bm{M}_{1}^{\mathrm{SAM}}(i) \cdot \bm{M}_{1'}^{\mathrm{SAM}}(j))}{Sum(\bm{M}_{1}^{\mathrm{SAM}}(i))}}
	\label{eq:3dm12}
\end{equation}  
\begin{equation}
	\bm{M}_{2 \rightarrow 1}(i,j) = {\frac{Sum(\bm{M}_{2}^{\mathrm{SAM}}(j) \cdot \bm{M}_{2'}^{\mathrm{SAM}}(i))}{Sum(\bm{M}_{2}^{\mathrm{SAM}}(j))}}
	\label{eq:3dm21}
\end{equation}  
where $1\leq i\leq N_1,1\leq j\leq N_2$, $Sum$ is the summation function of the binary matrix, $\bm{M}_{1 \rightarrow 2}(i,j)$ reveals the coverage rate of the $j$-th mask in $\bm{M}_{1'}^{\mathrm{SAM}}$ over the $i$-th mask in $\bm{M}_{1}^{\mathrm{SAM}}$.

By searching for the $n$-th row in $\bm{M}_{1 \rightarrow 2}$, the $m$-th column with the highest coverage rate can be found,  at this time, $(n, m)$ is considered as a preliminary match. Afterward, we also need to perform a check in $\bm{M}_{2 \rightarrow 1}$, where the maximum value of the $n$-th col in the $\bm{M}_{2 \rightarrow 1}$ should be taken at $m$-th row. This bidirectional verification prevents situations where the mask in one frame is too large and directly covers another.

\textbf{Matching Quality Checking:} Further quality checks are needed after obtaining the initial matching binary masks $\bm{M}_{1}^{\mathrm{SAM}}(m)$ and $\bm{M}_{2}^{\mathrm{SAM}}(n)$. In addition to the methods mentioned in the previous Sec.\ref{mask_check}, we also require that the coverage rate $\bm{M}_{1 \rightarrow 2}(m,n)$ and $\bm{M}_{2 \rightarrow 1}(m,n)$ should be greater than 0.95 to ensure the consistency and completeness of the matching mask content.

\textbf{Output:} We extract the corresponding foreground part from the input RGB images and optical flow label based on a matched binary mask, as shown in Fig.\ref{figfore3D}. The optical flow labels have already used $M_{RC}$, $M_{GC}$ and $M_{VSS}$ to remove incorrect areas.

\textbf{2D Foreground $F_{2d}$:} Based on the successful experience of previous researchers\cite{truong2023pdc,ling2024scaleraftcrossscalerecurrentallpairs}, we have also attempted to add 2D flight foreground, which, like the 3D flight foreground, is directly overlaid on the surface of the generated optical flow label (imagine a randomly textured piece of paper dancing in the air). For specific technical details, please refer to ScaleFlow++\cite{ling2024scaleraftcrossscalerecurrentallpairs}.

This pipeline can also render stereo 3D foreground pairs. Simply replace the input optical flow pairs with stereo pairs.
\subsection{Training of Optical Flow and Stereo}
In this section, we will introduce how to use the various components proposed in this article for training optical flow or stereo models.

After calculating the optical flow or stereo labels between two frames, we will combine masks and foreground modules according to the specific task type to filter the label results. Generally speaking, $M_{RC}$ and 3D flight foreground $F_{3d}$ have beneficial improvements for all types of scenes and tasks, especially when user-collected and application scenarios are similar. Secondly, we suggest that users try using $M_{VSS}$ and $M_{GC}$ when they need better extrapolation capabilities (as they will remove the inherent occlusion in the labels). The specific final label used is as follows:
 \begin{equation}
	f_{1\rightarrow 2}^{gt} = M_{RC}^{th1}M_{VSS}^{th2}M_{GC}^{th3}f_{1\rightarrow 2}
	\label{eq:eq166}
\end{equation}   
\begin{equation}
	M^{th}= \begin{cases}1 & \text { if } M <t h \text { and }  M_{valid}>0 \\ 0 & \text { otherwise }\end{cases}
\end{equation}
the points with $f_{1\rightarrow 2}^{gt}$ equal to 0 do not participate in the loss calculation. The binarization thresholds $th$ and effective regions $M_{valid}$ corresponding to different masks are shown in Tab.\ref{tab:t1}. We show examples of the complete SAG dataset in the supplementary materials; feel free to check them out.
\begin{table}[htbp]
	\centering
	\caption{\textbf{Binary threshold and effective region of mask.}}
	\setlength{\tabcolsep}{4.2pt}
	\begin{tabular}{lrrr}
		\toprule
		& \multicolumn{1}{l}{Binary $th$   (NeRF)} & \multicolumn{1}{l}{Binary  $th$ (3DGS)} & \multicolumn{1}{l}{$M_{valid}$} \\
		\midrule
		$M_{RC}$   &    $M_{RC}<0.4$     &   $M_{RC}<0.06$    & \multicolumn{1}{l}{Full image} \\
		$M_{occ}$  &       &       & \multicolumn{1}{l}{$M_{RC}$} \\
		$M_{VSS}$  &   $M_{VSS}<0.1$    &   $M_{VSS}<0.1$    &  \multicolumn{1}{l}{$M_{occ}$} \\
		$M_{GC}$   &   $M_{GC}<0.01$  &    $M_{GC}<0.01$   &  \multicolumn{1}{l}{$M_{occ}$} \\

		\bottomrule
	\end{tabular}%
	\label{tab:t1}%
\end{table}%

\section{Experimental} 
In this section, we first introduce the critical details of the dataset and experiments, and then further discuss the experimental results.
Specifically, the experiment consists of three aspects. Firstly, a comprehensive ablation of the proposed module was performed. Secondly, a quantitative comparison was made between the SAG framework and other advanced methods on public optical flow and stereo datasets. Lastly, we qualitatively evaluate the zero-shot generalization performance of SAG in real-world generalization scenarios.

\begin{table}[htbp]
	\centering
	\caption{\textbf{Training details of optical flow and stereo.} upper part of the table is the optical flow, and the lower part is the stereo.}
	\setlength{\tabcolsep}{2.4pt}
	\begin{tabular}{lccccc}
		\toprule
		\multicolumn{1}{c}{Training Data} & Method & Iteration & Size  & Batch & Lr \\
		\midrule
		GS15 & RAFT\cite{teed2020raft}  & 100k  & \multirow{6}[1]{*}{384$\times$768} & \multirow{7}[2]{*}{6} & \multirow{7}[2]{*}{0.00025} \\
		NeRF15 & RAFT  & 100k  &       &       &  \\
		GS58  & RAFT  & 200k  &       &       &  \\
		GS58  & Scale-flow\cite{ling2022scale} & 240k  &       &       &  \\
		GS58  & ScaleFlow++\cite{ling2024scaleraftcrossscalerecurrentallpairs} & 240k  &       &       &  \\
		S,T   & RAFT  & 100k  &       &       &  \\
		C     & RAFT  & 100k  & 384$\times$512 &       &  \\
		\midrule
		GS58  & RAFT-stereo\cite{lipson2021raft} & 300k  & \multirow{3}[2]{*}{384$\times$768} & \multirow{3}[2]{*}{4} & \multirow{3}[2]{*}{0.0002} \\
		GS58  & IGEV\cite{Xu_2023_CVPR}  & 300k  &       &       &  \\
		GS58+T+M  & IGEV  & 600k  &       &       &  \\
		GS15 & IGEV  & 100k  &       &       &  \\
		\bottomrule
	\end{tabular}%
	\label{tab:TRAINING}%
\end{table}%
\subsection{Generation Details}
\textbf{Generate Datasets:} We collected and constructed approximately 300 scenes, mainly indoor still life and outdoor courtyards. We use Zip-NeRF\cite{barron2023zip} and MIP-splatting\cite{Yu_2024_CVPR} for reconstruction separately. Zip-NeRF trained 30000 iterations with a batch size of 10240, and MIP-splatting also trained 30000 iterations. For each scene, we generate approximately 200 pairs of camera poses to generate optical flow and stereo datasets. The final optical flow and stereo datasets consist of approximately 58800 pairs of images (GS58, NeRF58). Moreover, for the convenience of ablation experiments, we extracted subsets from the complete GS58 and NeRF58 to form GS15 and NeRF15, with 15800 image pairs per subset.

\textbf{Generation 3D Foreground:} The 3D foreground we generate is derived from the reconstruction scene used in the previously generated dataset. Generally speaking, the scenes corresponding to GS58 can generate approximately 250000 pairs of effective flight foreground ($F_{3d}$), while those corresponding to GS15 can generate approximately 60500 pairs of effective flight foreground. During training, we enhanced the generated foreground data, including scaling, translation, color changes, etc. For specific details, please refer to RAFT\cite{teed2020raft} and RAFT-stereo\cite{lipson2021raft}.

\subsection{Training and Evaluation Details}
We mainly conduct quantitative comparisons of methods based on KITTI and Sintel, as they have publicly evaluated platforms and have been widely used in previous work. In addition, we also used the Quarter (Q) and Half (H) resolutions in Middlebury for stereo-matching evaluation (our method is trained on an approximate Quarter resolution).

\textbf{Datasets for Optical Flow Task:} 
200 image pairs from KITTI 2015 (K15)\cite{Menze2015ISA} , 194 image pairs from KITTI 2012 (K12)\cite{geiger2012we}, 3 static scenes from Middlebury\cite{baker2011database} (Midd-A). Around 300,000 KITTI raw data (Kraw) and 4000 multi-frame (K15m) images related to K15 were used for training in loss-driven self-supervised methods\cite{kong2022mdflow,luo2021upflow}. Moreover, commonly used pre-training datasets Flyingthings3D\cite{mayer2016large} (T), Flyingchairs2 (C2)\cite{mayer2016large}, Flyingchairs (C)\cite{mayer2016large}, Driving (D)\cite{mayer2016large}  and Sintel (S)\cite{butler2012naturalistic} have also been included in the comparison range.

\textbf{Datasets for Stereo Task:} 200 stereo pairs from KITTI 2015 (K15)\cite{Menze2015ISA} , 194 stereo pairs from KITTI 2012 (K12)\cite{geiger2012we}, 15 stereo pairs from Middlebury v3\cite{scharstein2014high} training set (Midd-T), the commonly used pre-training synthetic dataset SenceFlow\cite{mayer2016large} includes Flylingthings3D (T), Monkaa (M), and Driving (D).

\textbf{Training Details:}
Considering representativeness, we mainly choose RAFT\cite{teed2020raft} and IGEV\cite{Xu_2023_CVPR} as the main architectures for optical flow and stereo ablation. We also considered other methods, including Scale-flow\cite{ling2022scale}, ScaleFlow++\cite{ling2024scaleraftcrossscalerecurrentallpairs}, and RAFT-Stereo\cite{lipson2021raft}, to evaluate the effectiveness of the SAG Framework on various datasets. We list the specific training details for different methods and datasets in Tab.\ref{tab:TRAINING}.

During training, we directly overlay the generated flight foreground $F_{3d}$ onto the background data. Generally, the number of foreground overlays per frame does not exceed two. For more information, please refer to the supplementary materials.
\begin{figure}[!t]
	\centering
	\includegraphics[width=3.4in]{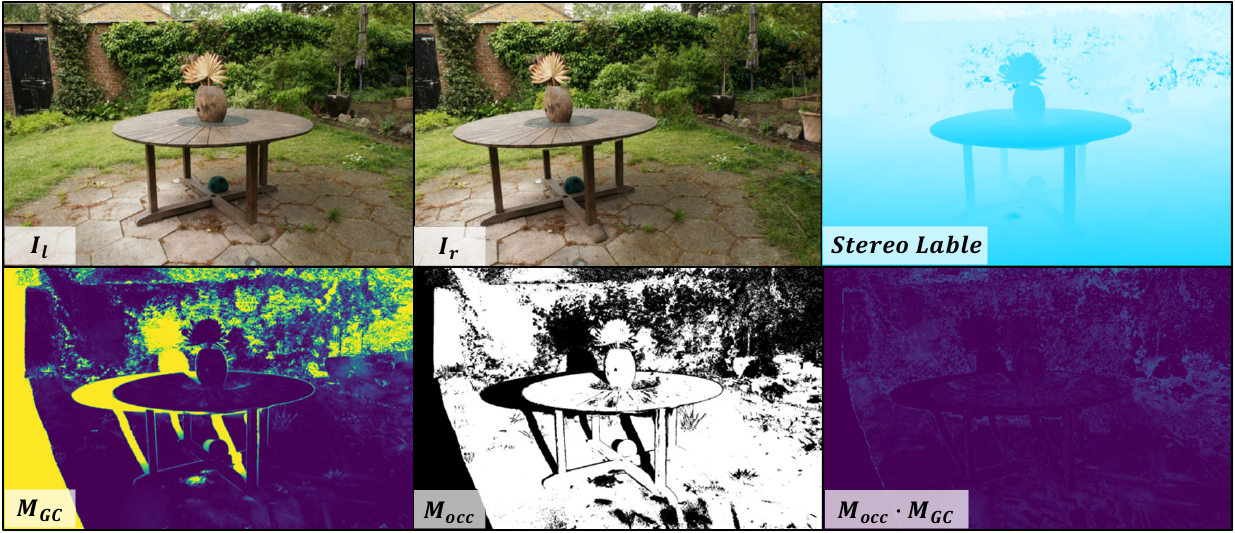}
		\caption{\textbf{Geometric Consistency Evaluation.} The generated stereo image pairs $I_l$ and $I_r$ used to evaluate are shown in the figure above. Because  $M_{GC}$ cannot evaluate the consistency of the occluded part, an occlusion mask needs to be used to eliminate it. }
	\label{fig_GCocc}
\end{figure}
\begin{table}[htbp]
	\centering
	\setlength{\tabcolsep}{4.8pt}
	\caption{\textbf{Geometric consistency evaluation for different rendering methods.} The indicator of median and mean depth is $GC_{l}$. We take the mean of each scene's image group as a representative.}
	\begin{tabular}{l|cc|cc|cc}
		\toprule
		\multicolumn{1}{r}{} & \multicolumn{2}{c}{ Median Depth $GC_{l}\downarrow$} & \multicolumn{2}{c}{ Mean Depth $GC_{l}\downarrow$} & \multicolumn{2}{c}{Time/s $\downarrow$} \\
		\midrule
		\multicolumn{1}{r}{} & 3DGS    & \multicolumn{1}{c}{NeRF } & 3DGS    & \multicolumn{1}{c}{NeRF } & 3DGS    & NeRF  \\
		\midrule
		bicycle & 0.27  & 3.71  & 0.69  & 3.96  & 1.38  & 19.03  \\
		bonsai & 0.17  & 2.29  & 0.74  & 2.51  & 1.79  & 6.90  \\
		counter & 0.12  & 4.53  & 0.28  & 4.67  & 1.71  & 6.90  \\
		flowers & 0.30  & 3.77  & 0.71  & 4.19  & 1.47  & 18.69  \\
		garden & 0.11  & 2.60  & 0.36  & 2.71  & 1.30  & 19.84  \\
		kichen & 0.09  & 3.29  & 0.24  & 3.39  & 1.22  & 6.87  \\
		room   & 0.11  & 6.07  & 0.26  & 6.14  & 1.12  & 6.91  \\
		treehill & 0.34  & 3.48  & 1.02  & 3.97  & 2.34  & 18.40  \\
		stump & 0.26  & 6.13  & 0.59  & 6.85  & 2.92  & 18.84  \\
		\bottomrule
	\end{tabular}%
	\label{tab:mm}%
\end{table}%

\subsection{Median Depth and Mean Depth} \label{mmu}
In this section, we use geometric consistency $M_{GC}$ as a quantitative indicator to evaluate the performance of different reconstruction baselines and depth rendering methods. Specifically, we generate a stereo-matching dataset (with an average disparity of about 50 pixels) on the commonly used reconstruction dataset 360v2\cite{barron2022mip}. Then, calculate the geometric consistency of the generated dataset. The calculation method for the evaluation metrics is as follows:
\begin{equation}
	GC_{l}= Mean(M_{occ} \cdot M_{GC})\cdot100
	\label{eq:gcloss}
\end{equation}   
where $M_{occ}$ represents the occlusion mask between two frames, as shown in Fig.\ref{fig_GCocc}, which is used to eliminate artifacts caused by occlusion, $Mean$ is a function used to calculate the mean.

The specific evaluation results are shown in Tab.\ref{tab:mm}. Firstly, it is evident that the geometric consistency of the 3DGS method is much better than that of the NeRF method, thanks to its explicit expression. Secondly, we also found that whether it is 3DGS or NeRF, the geometric consistency of median depth is always better than mean depth, which proves the rationality of using median depth instead of mean depth in this paper. Finally, the scene rendering time of 3DGS is only one-tenth that of NeRF, significantly improving efficiency while maintaining high data quality. This performance confirms the significance of expanding the 3DGS baseline.
\begin{table*}[htbp]
	\centering
	\caption{\textbf{Ablation Study for Optical Flow.} The best among the same group are bolded, and the second best is underlined. $EPE$ is the average end-to-end optical flow error, $Fl_{all}$ is the optical flow outlier rate (errors greater than 3 pixels or greater than 5\% are considered outliers). For details about $F_{2d}$ and $F_{3d}$, please refer to Sec.\ref{3Dfore}.}
	\setlength{\tabcolsep}{4.0pt}
	\begin{tabular}{cclcccccccccc}
		\toprule
		&       &       & \multicolumn{2}{c}{K15} & \multicolumn{2}{c}{K12} & \multicolumn{2}{c}{Midd-A} & \multicolumn{2}{c}{S-clean} & \multicolumn{2}{c}{S-final} \\
		\midrule
		\multicolumn{1}{l}{Method} & \multicolumn{1}{l}{Training Data} & Ablation & $EPE$   & $Fl_{all}$    & $EPE$   & $Fl_{all}$    & $EPE$   & $Fl_{all}$    & $EPE$   & $Fl_{all}$    & $EPE$   & $Fl_{all}$ \\
		\midrule
		\multirow{7}[2]{*}{RAFT\cite{teed2020raft}} & \multirow{7}[2]{*}{GS15} &       & 3.87  & 15.12  & 1.66  & 8.67  & 0.208  & 0.151  & 1.67  & 4.65  & 3.15  & 8.47  \\
		&       & $F_{2d}$   & 3.85  & 14.64  & 1.69  & 8.36  & 0.221  & 0.164  & 1.68  & 4.58  & 3.34  & 8.58  \\
		&       & $F_{2d}$+$M_{RC}$ & \textbf{3.53} & \textbf{12.17} & \textbf{1.52} & \underline{6.43}  & 0.213  & 0.165  & \underline{1.65}  & \underline{4.52}  & 3.22  & \underline{8.34}  \\
		&       & $F_{2d}$+$M_{VSS}$ & 4.10  & 16.07  & 1.75  & 9.18  & \textbf{0.190} & 0.166  & 1.70  & 4.61  & 3.14  & 8.44  \\
		&       & $F_{2d}$+$M_{GC}$ & 3.67  & 14.72  & 1.67  & 8.35  & 0.201  & 0.123  & 1.63  & 4.65  & \textbf{3.05} & 8.50  \\
		&       & $F_{2d}$+$M_{RC}$+$M_{GC}$ &\underline{3.59} & 13.15  & 1.53  & 6.60  & 0.199  & \underline{0.114}  & \textbf{1.61} & 4.66  & 3.10  & 8.40  \\
		&       & $F_{2d}$+$M_{RC}$+$M_{GC}$+$F_{3d}$ & 3.77  & \underline{13.13}  & \underline{1.53}  & \textbf{6.28} & \underline{0.192}  & \textbf{0.102} & 1.70  & \textbf{4.49} & \underline{3.08}  & \textbf{8.17} \\
		\midrule
		\multirow{7}[2]{*}{RAFT} & \multirow{7}[2]{*}{NeRF15} &       & 4.90  & 17.34  & 2.04  & 9.18  & 0.313  & 0.222  & 1.78  & 5.49  & 3.24  & 9.50  \\
		&       & $F_{2d}$   & 4.62  & 15.71  & 1.81  & 7.45  & 0.368  & 0.184  & 1.74  & 5.02  & 3.24  & 9.09  \\
		&       & $F_{2d}$+$M_{RC}$ & \underline{4.19}  & 14.83  & 1.76  & 8.12  & 0.300  & 0.231  & 1.73  & 5.03  & 3.29  & 8.96  \\
		&       & $F_{2d}$+$M_{VSS}$ & 4.40  & 15.70  & \underline{1.72}  & 7.76  & 0.257  & \underline{0.129}  & 1.72  & 5.05  & \underline{3.23}  & 8.96  \\
		&       & $F_{2d}$+$M_{GC}$ & 4.42  & 18.28  & 1.77  & 9.59  & \textbf{0.201 } & 0.186  & 1.82  & 5.16  & 3.31  & 9.09  \\
		&       & $F_{2d}$+$M_{GC}$+$M_{VSS}$+$M_{RC}$ & \textbf{4.19} & \textbf{14.15} & \textbf{1.62} & \textbf{6.80} & 0.280  & 0.175  & \underline{1.73}  & \underline{4.96}  & 3.33  & \underline{8.81}  \\
		&       & $F_{2d}$+$M_{GC}$+$M_{VSS}$+$M_{RC}$+$F_{3d}$ & 4.35  & \underline{14.55}  & 1.74  & \underline{7.01}  & \underline{0.203}  & \textbf{0.114} & \textbf{1.64} & \textbf{4.51} & \textbf{3.00} & \textbf{8.31} \\
		\midrule
		\multirow{5}[2]{*}{RAFT} & S &       & 8.50  & 18.31  & 2.38  & 8.01  & 0.230  & 0.220  & -  & -  & -  & -  \\
		& T &       & 8.20  & 23.42  & 3.14  & 14.16  & 0.240  & 0.170  & 1.79  & 5.37  & 3.04  & 9.25  \\
		& T & $F_{3d}$   &  7.59  &  27.35  &  3.38 &  15.38 &  0.233  &  0.182  & 1.64  &  4.80 & 2.98  & 8.40\\
		&C2 &       & 10.87  & 36.86  & 4.20  & 26.00  & 0.460  & 0.790  & 2.39  & 7.08  & 4.04  & 11.15  \\
		& C2 & $F_{3d}$   & 8.04  & 27.87  & 3.35  & 17.26  & 0.379  & 0.459  & 1.99  & 5.86  & 3.66  & 9.98  \\
		\midrule
		RAFT\cite{teed2020raft}  & GS58  &  $F_{2d}$+$M_{RC}$+$M_{GC}$+$F_{3d}$ & 3.55  & 11.89  & 1.52  & 6.01  & 0.177  & 0.115  & 1.57  & 4.23  & 3.13  & 7.95  \\
		Scale-flow\cite{ling2022scale} & GS58  &  $F_{2d}$+$M_{RC}$+$M_{GC}$+$F_{3d}$ & 3.34  & 11.77  & 1.42  & 5.58  & 0.170  & 0.117  & 1.46  & 4.11  & 3.02  & 7.95  \\
		ScaleFlow++\cite{ling2024scaleraftcrossscalerecurrentallpairs} & GS58  &  $F_{2d}$+$M_{RC}$+$M_{GC}$+$F_{3d}$ & 2.90  & 9.56  & 1.37  & 6.20  & 0.166  & 0.135  & 1.38  & 3.81  & 3.03  & 7.53  \\
		\bottomrule
	\end{tabular}%
	\label{tab:addlabel}%
\end{table*}%

\subsection{Ablation Study for Optical Flow}
In this section, we investigated the effectiveness of the various components of the generalization framework SAG proposed in this paper in the optical flow task. We also compared the SAG with commonly used synthetic datasets.

\textbf{Mask Analysis of Optical Flow:} \label{mmd}
Tab.\ref{tab:addlabel} shows the results of RAFT trained with different masks. Firstly, we found that in most cases, the training effect always improves after using masks. Specifically, $M_{RC}$ showed the most significant improvement among all masks, demonstrating the effectiveness of our motivation to evaluate dataset quality from a probabilistic perspective. Secondly, most indicators achieved the best results after using all masks,  indicating that the proposed masks are compatible.

\textbf{Comparison with Synthetic Datasets:}
We compared NeRF15 and GS15 with commonly used pre-training datasets S, T, and C2 in the optical flow task, and all datasets were trained from scratch using almost the same configuration based on RAFT. The specific training configurations are detailed in Tab.\ref{tab:TRAINING}.
The comparison results are shown in Tab.\ref{tab:addlabel}. The performance of NeRF15 and GS15 generated by SAG is always ahead of the synthetic data set, especially in the test of real scenarios (KITTI, Midd-A). This lead is mainly due to the fact that the data generated by SAG is closer to the real scene while maintaining high quality. It turns out that SAG can be used as a replacement and complement to existing synthetic datasets in optical flow tasks for real-world generalization or model pre-training.

\textbf{NeRF and 3DGS Baseline:}
We compared the performance of Zip-NeRF (NeRF) and Mip-Splatting (3DGS) as generating baselines in Tab.\ref{tab:addlabel}. The results showed that 3DGS as the generating baseline was significantly better than NeRF, consistent with the experimental conclusions we obtained in Tab.\ref{tab:mm}. In addition, the data generation speed of the 3DGS baseline (with an average 10-fold increase in speed) is much higher than NeRF, so in subsequent experiments, we mainly rely on 3DGS. It should be noted that NeRF is not without its advantages, as it has accumulated many unique technological advantages in many special application scenarios, such as fuzzy input, light field reconstruction, and reflective surfaces.

\textbf{3D Flight Foreground}
As shown in Tab.\ref{tab:addlabel}, using 3D foreground $F_{3d}$ always brings performance improvements to the method. We found that the improvement in the synthetic dataset Sintel is more significant than in KITTI, mainly due to the higher proportion of foreground components in Sintel. In addition, we also applied $F_{3d}$ to the synthetic datasets T and C, comprehensively improving their generalization performance, proving that the 3D foreground generation module can be directly applied as a separate generalization enhancement method to existing training systems.

\textbf{Number of Scenes and Stronger Baseline:}
We also attempted to train with more scenes. As expected, compared to GS15, GS58 has improved in all geometric metrics, which is a crucial advantage of our work: SAG can quickly expand the scene captured by hand to the dataset, thereby improving the performance of the method. Moreover, we tested other more complex optical flow baselines, such as Scale-flow and ScaleFlow++, which performed better than RAFT, indicating that our data-driven approach can be applied to complex baselines.

\begin{table*}[htbp]
	\centering
	\setlength{\tabcolsep}{4.8pt}
	\caption{\textbf{Evaluation of Optical Flow Generalization Performance.} The best among the same group are bolded,and the second best is underlined.}
	\begin{tabular}{cccccccccc}
		\toprule
		&       &       & \multicolumn{2}{c}{K15} & \multicolumn{1}{c}{K15-test} & \multicolumn{2}{c}{K12} & \multicolumn{2}{c}{S-test} \\
		\midrule
		Training type & Method & Training data & $EPE$   & $Fl_{all}$  & \multicolumn{1}{c}{$Fl_{all}$} & $EPE$    & $Fl_{all}$  & $EPE$(clean) & $EPE$(final) \\
		\midrule
		\multicolumn{1}{c}{\multirow{4}[2]{*}{Supervised Generalization }} & LiteFlowNet2\cite{hui2020lightweight} & C + T & 8.97  & 25.9  &    -   & 3.42  &   -    & 3.96  & 6.02 \\
		& VCN\cite{yang2019volumetric}   & C + T & 8.36  & 25.1  &    -   &    -   &    -   &    -   & - \\
		& RAFT\cite{teed2020raft}   & C + T & 5.04  & 17.4  &    -   &    -   &   -    &    -   & - \\
		& SEA-RAFT(L)\cite{wang2024sea} & C + T & 3.62 & 12.9 &   -    &     -  &    -   &    -   &  -\\
		& GMFlow\cite{xu2022gmflow} & C + T & 7.47 & 23.40 &   -    &     3.42  &    14.93   &    -   &  -\\
		\midrule
		\multicolumn{1}{c}{\multirow{5}[2]{*}{Self-supervised Fine-tuning}} & UFlow\cite{jonschkowski2020matters} & K15m+Kraw & 2.71  &    -   & \multicolumn{1}{c}{11.13} & 1.68  &   -   & 5.21  & 6.5 \\
		& MDFlow-fast\cite{kong2022mdflow} & C+S/K15m & 3.02  &    -   & \multicolumn{1}{c}{11.43} &   -    &    -   & 4.73  & 5.99 \\
		& CoT-AMFlow\cite{pmlr-v155-wang21a} & K15m+S &    -   &    -   &    -   &    -   &   -    & 3.96  & 5.14 \\
		& UPFlow\cite{luo2021upflow} & K15m+S & 2.45 &   -    & \multicolumn{1}{c}{9.38} & 1.27 &    -   & 4.68  & 5.32 \\
		& SMMSF\cite{hur2021self} & K15m+Kraw & 6.04  & 18.81 & \multicolumn{1}{c}{15.97} &    -   &   -    &   -   &  -\\
		\midrule
		\multicolumn{1}{c}{\multirow{11}[2]{*}{Self-supervised Generalization}} & COTR\cite{jiang2021cotr}  & MegaDepth & 6.12  & 16.9  &    -   & 2.26  & 10.5  &    -   &  \\
		& GLU-Net\cite{truong2020glu} & COCO  & 7.49  & 33.83 &   -    & 3.14  & 19.76 &   -    & - \\
		& PDC-Net+\cite{truong2023pdc} & COCO  & 4.53  & 12.62 &   -    & 1.76  & 6.6   &   -    & - \\
		& PDC-Net\cite{truong2021learning} & COCO  & 5.22  & 15.13 &    -   & 2.08  & 7.98  &    -   &  -\\
		& MDFlow-fast\cite{kong2022mdflow} & S     & 10.05 & 23.12 &  -    & 3.49  & 12.17 &   -    & - \\
		& MDFlow-fast\cite{kong2022mdflow} & GTA5  & 9.13  & 25.01 &   -    & 3.85  & 14.33 &   -    & - \\
		& Scale-flow\cite{ling2022scale}  & ADF58\cite{Ling_2024_CVPR} & 3.88  & 13.36 & \multicolumn{1}{c}{13.47} & 1.59  & 6.97  & 2.57  & 4.65 \\
		& RAFT\cite{teed2020raft}   &  ADF58\cite{Ling_2024_CVPR}  & 4.17  & 13.9  & \multicolumn{1}{c}{13.41} & 1.59  & 6.43  &    -   &  -\\
		& Scale-flow\cite{ling2022scale} & GS58 (ours) & \underline{3.34} & \underline{11.77}  &    \underline{11.31}   & \underline{1.42} & \textbf{5.58} & \underline{2.35}      &  \underline{4.60}\\
		& RAFT\cite{teed2020raft}  & GS58 (ours) & 3.55  & 11.89  &   -    & 1.52  & \underline{6.01}  &     -  &  -\\
		& ScaleFlow++\cite{ling2024scaleraftcrossscalerecurrentallpairs} & GS58 (ours) & \textbf{2.90}  & \textbf{9.56} &    \textbf{9.73}   & \textbf{1.37}  & 6.20  & \textbf{2.04} & \textbf{4.26} \\
		\bottomrule
	\end{tabular}%
	\label{tab:opticalflow}%
\end{table*}%

\begin{figure}[!t]
	\centering
	\includegraphics[width=3.45in]{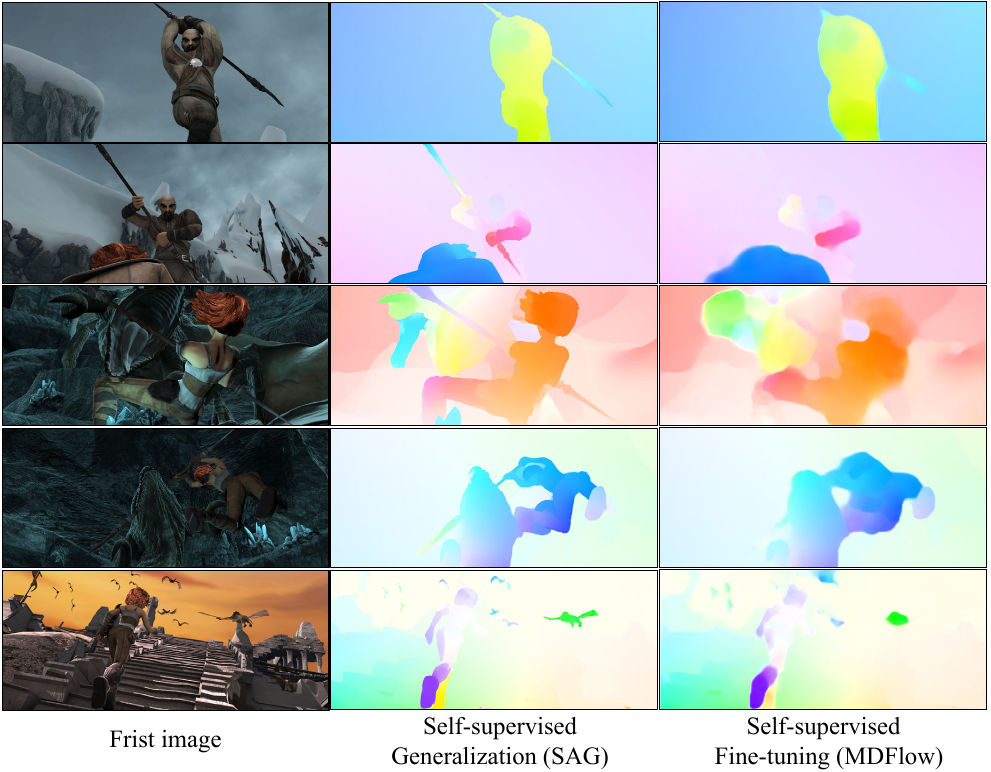}
	\caption{\textbf{Visual Comparison of Zero-shot Generalization on Sintel Test Set.} Our data-driven self-supervised methods have demonstrated cross-generational progress, with zero-shot generalization performance on Sintel far surpassing loss-driven self-supervised fine-tuning methods. Especially in small objects (poles, flying birds) and in occlusion areas, our solution has unprecedented clarity.
	}
	\label{fig_flows}
\end{figure}
\subsection{Evaluation of Optical Flow Generalization Performance}
This section compares SAG with several state-of-the-art optical flow generalization methods. In addition to conducting performance evaluations on K15 and K12 with existing truth values, we also conducted performance evaluations on the official testing platforms of KITTI and Sintel.

\textbf{Self-supervised Generalization:}  As shown in Tab.\ref{tab:opticalflow}, our GS58 achieved the best performance among all self-supervised generalization methods. The performance surpasses the previous state-of-the-art PDC-Net+ (TPAMI 2023). In addition, compared with other self-supervised methods, our method is suitable for various complex optical flow frameworks and has better universality.

\textbf{Supervision and Fine-tuning:}
We also compared our method with existing supervised generalization and self-supervised fine-tuning methods. Our SAG is in a leading position compared to the current state-of-the-art supervised generalization method SEA-RAFT. Even compared with the method of self-supervised fine-tuning, our method achieved highly competitive performance on KITTI and comprehensively surpassed other self-supervised methods on Sintel (2.04 v.s. 3.96). As shown in Fig.\ref{fig_flows}, SAG can predict details better and cope with occlusion well. These performances demonstrate the excellent generalization ability of our data-driven training framework.

\begin{table*}[htbp]
	\centering
	\caption{\textbf{Ablation Study for Stereo.} $ALL$ and $NOC$ are the stereo outlier rates for all areas and no-occlusion areas, respectively. In KITTI,  errors greater than 3 pixels or greater than 5\% are considered outliers, while in Middle, errors greater than 2 pixels are considered outliers.}
	\begin{tabular}{cclcccccccc}
		\toprule
		&       &       & \multicolumn{2}{c}{K15} & \multicolumn{2}{c}{K12} & \multicolumn{2}{c}{Midd-T(H)} & \multicolumn{2}{c}{Midd-T(Q)} \\
		\midrule
		Method & \multicolumn{1}{l}{Training Data} & Ablation & $ALL$   & $NOC$   & $ALL$   & $NOC$   & $ALL$   & $NOC$   & $ALL$   & $NOC$ \\
		\midrule
		\multirow{8}[2]{*}{IGEV\cite{Xu_2023_CVPR}} & \multirow{8}[2]{*}{GS15} &       & 6.69  & 6.43  & 6.92  & 6.15  & 12.99  & 9.66  & 9.39  & 6.65  \\
		&       & $M_{RC}$   & 4.72  & 4.42  & 4.04  & \textbf{3.34} & \underline{12.77}  & 9.07  & \underline{9.52}  & 6.50  \\
		&       & $M_{VSS}$  & 6.24  & 6.05  & 6.10  & 5.51  & 13.24  & 9.60  & 10.20  & 6.97  \\
		&       & $M_{GC}$   & 7.48  & 7.23  & 7.03  & 6.38  & 14.78  & 9.35  & 11.30  & 6.33  \\
		&       & $M_{RC}$+$F_{3d}$ & \underline{4.67}  & \underline{4.39}  & \underline{4.02}  & 3.41  & \textbf{12.28} & \underline{8.89}  & \textbf{8.48} & \textbf{5.69} \\
		&       & $M_{GC}$+$M_{RC}$+$F_{3d}$ & 4.69  & 4.41  & 4.43  & 3.77  & 13.51  & 8.97  & 10.37  & 6.18  \\
		&       & $F_{2d}$+$M_{GC}$+$M_{VSS}$+$M_{RC}$+$F_{3d}$ & 4.75  & 4.46  & 4.06  & 3.52  & 13.18  & 8.57  & 10.56  & 6.22  \\
		&       & $F_{2d}$+$M_{GC}$+$M_{RC}$+$F_{3d}$ & \textbf{4.61} & \textbf{4.34} & \textbf{3.96} & \underline{3.40}  & 12.83  & \textbf{8.40} & 10.07  & \underline{6.09}  \\
		\midrule
		RAFT-stereo\cite{lipson2021raft} & GS58  & $F_{2d}$+$M_{GC}$+$M_{RC}$+$F_{3d}$ & 4.60  & 4.35  & 4.40  & 3.61  & 15.31  & 10.58  & 9.97  & 5.70  \\
		IGEV\cite{Xu_2023_CVPR}  & GS58  & $F_{2d}$+$M_{GC}$+$M_{RC}$+$F_{3d}$ & 4.40 & 4.16 &4.09 & 3.47 & 12.20  & 7.94  & 8.64  & 4.89  \\
		IGEV\cite{Xu_2023_CVPR}  & GS58+T+M  & $F_{2d}$+$M_{GC}$+$M_{RC}$+$F_{3d}$ & 4.09   & 3.84  &   3.15    &    2.70   & 9.17 & 6.26  & 7.14 & 4.61  \\
		\bottomrule
	\end{tabular}%
	\label{tab:stereoab}%
\end{table*}%

\begin{table*}[htbp]
	\centering
	\setlength{\tabcolsep}{4.8pt}
	\caption{\textbf{Evaluation of Stereo Generalization Performance.} The best among the same group are bolded, and the second best is underlined. Due to the complete source code of Ns-RAFT-Stereo not being publicly available, the results reported in the table are our reproduction of the original paper.}
	\begin{tabular}{ccccccccccc}
		\toprule
		&       &       & \multicolumn{2}{c}{K15} & \multicolumn{2}{c}{K12} & \multicolumn{2}{c}{Midd-T(H)} & \multicolumn{2}{c}{Midd-T(Q)} \\
		\midrule
		Training type & Method & Training data & $ALL$   & $NOC$   & $ALL$   & $NOC$   & $ALL$   & $NOC$   & $ALL$   & $NOC$ \\
		\midrule
		\multicolumn{1}{c}{\multirow{4}[2]{*}{Self-supervised Generalization}} & NS-RAFT-Stereo\cite{tosi2023nerf} & NS65 & 6.33  & 6.04  & 4.64  & 3.95  & \underline{14.25}  & \underline{10.25}  & 12.56  & 8.67  \\
		& MfS-PSMNet\cite{watson-2020-stereo-from-mono} & COCO+DIW+ADE20K   & 5.18  & 4.91  &    -   &    -   & 17.56 & 13.45 & 12.07 & 9.09 \\
		& RAFT-stereo\cite{lipson2021raft} & GS58 (ours) & \underline{4.60}  & \underline{4.35}  & \underline{4.40}  & \underline{3.61}  & 15.31  & 10.58  & \underline{9.97}  & \underline{5.70} \\
		& IGEV\cite{Xu_2023_CVPR}  & GS58 (ours) & \textbf{4.40} & \textbf{4.16} & \textbf{4.09} & \textbf{3.47} & \textbf{12.20}  & \textbf{7.94}  & \textbf{8.64}  & \textbf{4.89}  \\
		\midrule
		\multicolumn{1}{c}{\multirow{7}[2]{*}{Supervised Generalization}} & RAFT-stereo\cite{lipson2021raft} & \multicolumn{1}{c}{\multirow{7}[2]{*}{T+M+D}} & 5.45  & 5.44  & 4.35  & \underline{3.86}  & 11.21  & 8.66  & 10.25  & 7.44  \\
		& IGEV\cite{Xu_2023_CVPR}  &       & 6.04  & 5.77  & 5.16  & 4.52  & \underline{10.07} & \underline{7.24} & \underline{8.97} & \underline{6.31}  \\
		& ITSA-CFNet\cite{chuah2022itsa} &       & \underline{4.96}  & \underline{4.76}  & \underline{4.20}  &   -    & 18.01  & 14.00  & 13.32  & 9.73  \\
		& CREStereo\cite{li2022practical} &       & 5.79  & 5.40  &   -    &    -   & 17.57  & 13.87  & 12.88  & 8.85  \\
		& ITSA-GWCNet\cite{chuah2022itsa} &       & 5.6   & 5.39  & 4.90   &    -   & 19.38 & 15.95 & 14.36 & 10.76 \\
		& SGM + NDR\cite{aleotti2021neural} &       & 5.41  & 5.12  & 6.00     & 5.00     & 17.7  & 13.51 & 11.75 & 7.93 \\
		& DSMNet\cite{zhang2020domain} &       & 5.5   & 5.19  &   -    &    -   & 26.75 & 21.8  & 15.52 & 11.49 \\
		& IGEV\cite{Xu_2023_CVPR} &  GS58+T+M (ours)   & \textbf{4.09}   & \textbf{3.84}  &   \textbf{3.15}    &    \textbf{2.70}  & \textbf{9.17} & \textbf{6.26}  & \textbf{7.14} & \textbf{4.61} \\
		\bottomrule
	\end{tabular}%
	\label{tab:stereo}%
\end{table*}%

\subsection{Ablation Study for Stereo}
In this section, we further explore the effectiveness of the main components of SAG in the stereo task.

\textbf{Mask Analysis of Stereo:}
Tab.\ref{tab:stereoab} shows the effectiveness of training IGEV using different masks. Similar to optical flow, in most cases, the performance always improves after using masks, especially $M_{RC}$ masks. Specifically, we found that using only $M_{RC}+F_{3d}$ resulted in better performance in the Middle dataset. This is easy to understand, as $M_{RC}$ does not exclude occluded parts, while part of GS15 is composed of daily scenes similar to Middlebury, and the inference learning of occluded parts can be better generalized in similar scenes.

\textbf{3D Flight Foreground:}
We further validated the effectiveness of the 3D foreground, observing $M_{RC}$ and $M_{RC}+F_{3d}$ in Tab.\ref{tab:stereoab}. After using the 3D foreground, the model's performance was significantly improved, especially in predicting occluded parts. This improvement validates our motivation to use 3D foreground enhancement networks for learning occluded parts.

\textbf{Mixed Training:}
We also attempted to mix the commonly used pre-training datasets T and M with GS58 for training. As shown in the last row of Tab.\ref{tab:stereoab}, the generalization performance of the method has been comprehensively improved, proving that our SAG and existing pre-training methods are not conflicting, and can be combined to further improve the generalization performance of deep models.
\begin{figure}[!t]
	\centering
	\includegraphics[width=3.4in]{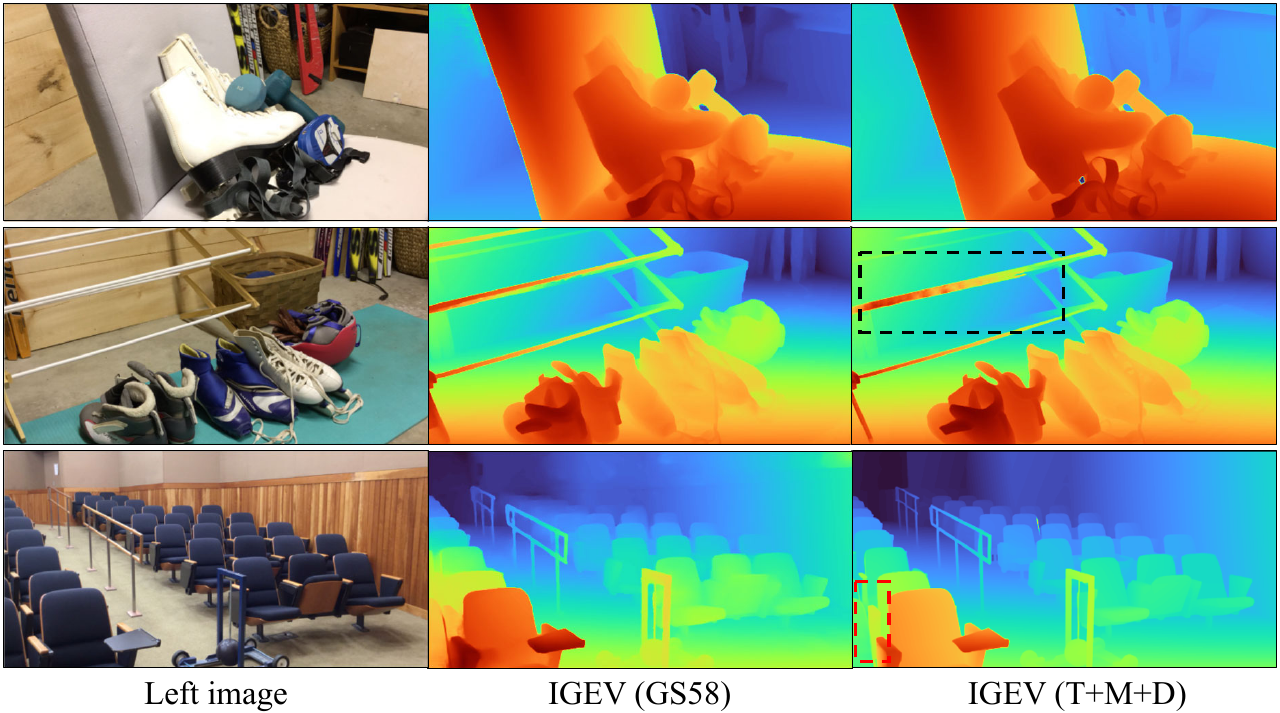}
	\caption{\textbf{Visual Comparison of Zero-shot Generalization on Middlebury.} Our data-driven self-supervised framework demonstrates better generalization performance than existing synthetic data generalization schemes. This is thanks to SAG's ability to quickly and cost-effectively customize real-world datasets to obtain sufficient prior knowledge to predict occlusion (chair back in the red box) and small objects (pole in the black box).}
	\label{fig_stereog}
\end{figure}

\subsection{Evaluation of Stereo Generalization Performance}
This section quantitatively compares SAG with state-of-the-art self-supervised generalization methods and  supervised generalization methods.

By observing the IGEV and RAFT Stereo in Tab.\ref{tab:stereo}, our scheme's generalization performance is on par with or surpasses the commonly used scene-flow datasets (T+M+D) in most metrics. This outstanding performance demonstrates the potential of our SAG framework to replace traditional synthetic datasets as a high-performance, low-cost alternative.
Secondly, compared with similar self-supervised generalization methods, our scheme has apparent advantages, surpassing most existing ones, especially on the KITTI dataset (with a significant deviation from the training set domain), reflecting the superiority of SAG generalization ability.

\subsection{Zero-shot Generalization in Real Word}
In addition to the quantitative evaluation in the previous chapters, we also conducted a visual qualitative evaluation, mainly based on the Davis dataset and some daily scenes collected by our mobile phones, presented in the form of videos. Welcome to check them on our code homepage\footnote{\url{https://github.com/HanLingsgjk/UnifiedGeneralization}}.

We have selected some typical examples for presentation. As shown in Fig.\ref{fig_stereog}, our SAG framework performs better than previous solutions based on synthetic datasets. Thanks to the prior knowledge learned from a large amount of input RGB images, it can better handle occlusion and small objects. With the continuous expansion and filtering of the RGB image set, SAG will demonstrate better performance.

\begin{figure*}[!t]
	\centering
	\includegraphics[width=7in]{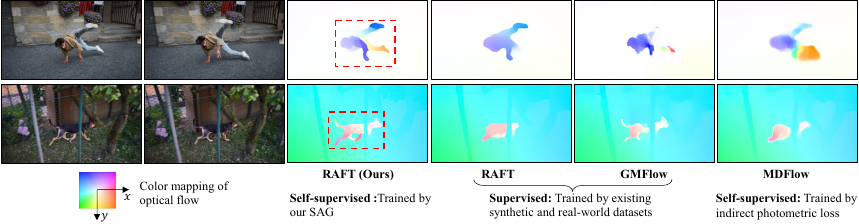}
	\caption{\textbf{Visual Comparison of Zero-shot Generalization Performance of Optical Flow in Natural Scenes.} SAG exhibits excellent generalization effects on unseen dynamic foregrounds (street dancer, dogs) in real-world scenarios. Compared to previous supervised and self-supervised generalizations, SAG can handle occlusions and predict details better.}
	\label{fig_opticalg}
\end{figure*}

Compared to the stereo task, SAG has greater application significance in the field of optical flow, as real-world labels for optical flow are more challenging to obtain. As shown in Fig.\ref{fig_opticalg}, compared with the previous loss-driven self-supervised method, the results of data-driven SAG are clearer and more accurate. Compared to supervised training-based methods, our approach has fewer artifacts. Because most current optical flow datasets are synthetic, artifacts may inevitably occur in real-world applications due to domain bias. On the contrary, our SAG can construct approximate datasets based on real-world data, learn sufficient prior knowledge, and reduce artifacts.

\begin{figure}[!t]
	\centering
	\includegraphics[width=3.4in]{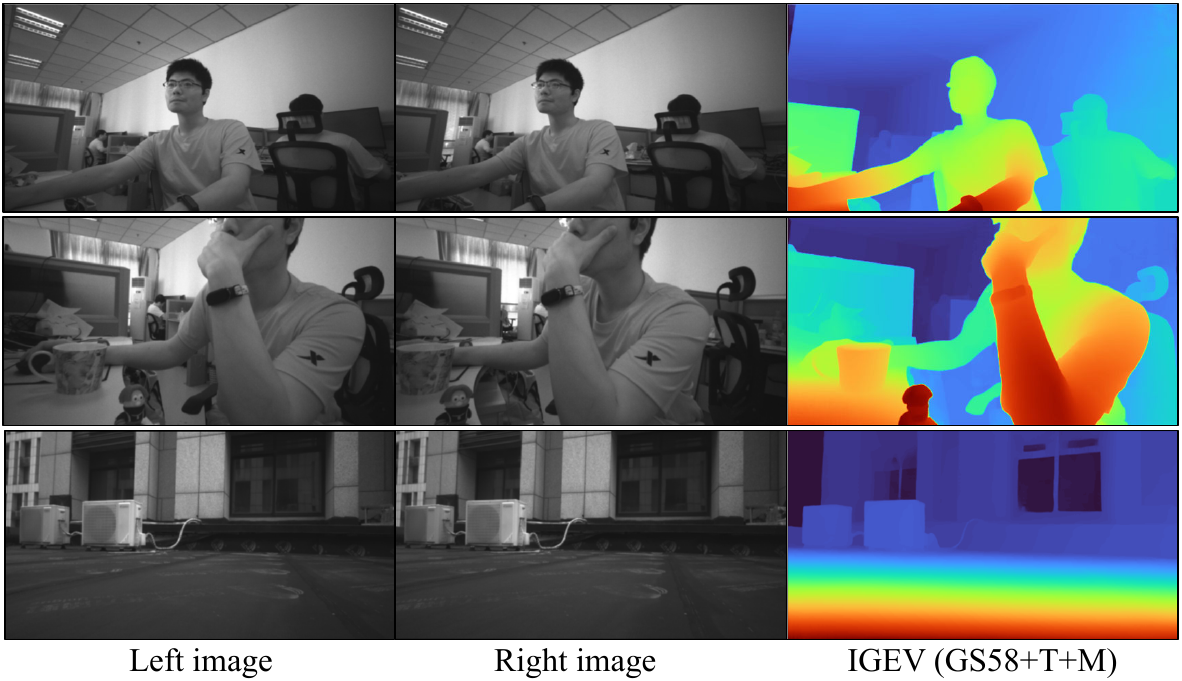}
	\caption{\textbf{Generalization Results on A Real Stereo camera.} The author of this paper, Han Ling, used an Intel D455 stereo camera to capture himself and the roof outside the laboratory. It can be seen that SAG still has good generalization performance in new devices and scenarios.}
	\label{fig_455}
\end{figure}

\section{Limitation} 
As shown in Fig.\ref{fig_455}, the reconstruction method used in this article cannot reconstruct reflective surfaces such as glass, resulting in void artifacts at the window. In addition, our current reconstruction scenarios are limited to static scenes.  These limitations hinder SAG from building generalized datasets in driving scenarios. Fortunately, our SAG framework can continue to benefit from the rapid development of NeRF and 3DGS methods. At present, the above restrictions have been partially resolved\cite{verbin2022ref,attal2021torf,luiten2023dynamic,yang2023emernerf,wu2022d}.

\section{Conclusion} 
We propose a unified self-supervised training framework, SAG, for two different motion and 3D perception tasks: optical flow and stereo. Unlike previous loss-driven self-supervised methods, SAG is data-driven for self-supervised learning of 3D structure and motion, which utilizes advanced reconstruction techniques to extract 3D structural information from RGB images and generate a controllable dataset for training. In order to overcome the inevitable defects in the generation process, we proposed the first universal reconstruction quality evaluation index from a probabilistic perspective, reconstruction confidence. We proved its effectiveness through a large number of experiments. We have also designed a 3D flight foreground automatic rendering pipeline to address the issue of insufficient motion foreground in generated data. The final SAG trained model achieved state-of-the-art or highly competitive performance on the current mainstream optical flow/stereo datasets, while having much lower training costs and better universality than previous training methods.

A key contribution of this paper is providing a data-driven self-supervised training scheme. Because it does not involve interference with the method itself or the loss function, it can be applied to the training of more complex deep networks, greatly expanding the scope of framework applications. We hope the work presented in this article can help many excellent optical flow and stereo models perform better in the real world, thereby serving production and daily life.

\bibliographystyle{IEEEtran}
\bibliography{main}
\end{document}